# Long-Term Prediction Accuracy Improvement of Data-Driven Medium-Range Global Weather Forecast

Yifan Hu[1, 2], Fukang Yin[2, *], Weimin Zhang[2, *], Kaijun Ren[2], Junqiang Song[2], Kefeng Deng[2], and Di Zhang[2]

[1]College of Computer Science and Technology, National University of Defense Technology, Changsha, P.R. China, 410073

[2]College of Meteorology and Oceanography, National University of Defense Technology, Changsha, P.R. China, 410073

**Abstract:** Long-term stability stands as a crucial requirement in data-driven medium-range global weather forecasting. Spectral bias is recognized as the primary contributor to instabilities, as data-driven methods difficult to learn small-scale dynamics. In this paper, we reveal that the universal mechanism for these instabilities is not only related to spectral bias but also to distortions brought by processing spherical data using conventional convolution. These distortions lead to a rapid amplification of errors over successive long-term iterations, resulting in a significant decline in forecast accuracy. To address this issue, a universal neural operator called the Spherical Harmonic Neural Operator (SHNO) is introduced to improve long-term iterative forecasts. SHNO uses the spherical harmonic basis to mitigate distortions for spherical data and uses gated residual spectral attention (GRSA) to correct spectral bias caused by spurious correlations across different scales. The effectiveness and merit of the proposed method have been validated through its application for spherical Shallow Water Equations (SWEs) and medium-range global weather forecasting. Our findings highlight the benefits and potential of SHNO to improve the accuracy of long-term prediction.

## 1 Introduction

Accurate and timely weather forecasts play an important role in many aspects of human society. In the past few years, numerical weather prediction (NWP) has been the most commonly used tool for weather forecasting (L. Chen et al., 2023; Lam et al., 2023), which simulates the future state of the atmosphere by solving the partial differential equations (PDEs) numerically (Bauer et al., 2015). Although NWP models can get accurate forecasts, they are often slow and need the support of high-performance computing systems (Bauer et al., 2015; Bi et al., 2023; L. Chen et al., 2023; Lam et al., 2023). Moreover, errors in initial conditions, approximations of physical processes in parameterizations, and the chaos of the atmosphere introduce uncertainties to NWP (Bauer et al., 2015; L. Chen et al., 2023).

Recently, deep learning has revolutionized the field of weather forecasts for obtaining more timely forecasts and more accurate results. For example, Rasp and Thuerey (2021) used a deep residual convolutional neural network (CNN) known as ResNet (He et al., 2016) to do continuous forecasts at a spatial resolution of $5.625° \times 5.625°$ and obtain similar performance compared to a physical baseline at a similar resolution. FourCastNet (Pathak et al., 2022) firstly improved the data-driven global weather forecast model's resolution to $0.25° \times 0.25°$, but the forecasting accuracy is slightly below the most advance NWP i.e. the operational integrated forecasting system (IFS) of the European Centre for Medium-Range Weather Forecasts (ECMWF). Before long, data-driven weather forecasting system achieved new breakthroughs. For example, Pangu-Weather (Bi et al., 2023) produces stronger deterministic forecast results than the operational IFS on all tested weather variables against reanalysis data. Soon after, GraphCast (Lam et al., 2023) achieved better results than IFS on more variables and support better severe event prediction. In 2023, a vision transformer variant called FengWu (K. Chen et al., 2023) solves the medium-range forecast problem from a multi-modal and multi-task perspective, and achieves state-of-the-art for longer forecast lead times. Moreover, FuXi (L. Chen et al., 2023) published with the comparable performance to ECMWF ensemble mean (EM) in 15-day forecasts.

However, conventional convolution and Transformer models ignore the fact that the data is on the sphere, which would introduce distortions. These distortions seriously affect the performance of iterative forecasts. To reduce the accumulation errors for longer effective forecasts, Pangu-Weather (Bi et al., 2023) trained the model on 4 different lead times and used a greedy hierarchical temporal aggregation strategy to minimize the number of iteration steps. Similarly, to optimize performance for both short and long lead times, Fuxi (L. Chen et al., 2023) used a cascade (Ho et al., 2022; Li et al., 2015) model architecture and fine-tuned the pre-trained models in specific 5-day forecast time windows. FengWu (K. Chen et al., 2023) proposed the replay buffer to store the predicted results from previous optimization iterations and used them as the current model's input, miming the intermediate input error during the auto-regressive inference stage. Although these methods have achieved good results, they still suffer distortions for the spherical data. To settle these distortions, Weyn et al. (2020) introduced cubed-sphere remapping to minimize the distortion on the cube faces and provide natural boundary conditions for padding in the convolution operations. Shen et al. (2021) analyzed the equivariance error in the spherical domain for neural networks theoretically and designed a spherical equivariant CNN to settle the distortions from projection and the ineffective translation equivariance. McCabe et al. (2023) used the Double Fourier Sphere (DFS) method to correct the artificial discontinuity induced by the 2D Fast Fourier Transform (FFT), leads to significantly lower errors in long-range forecasts. But DFS still introduces spatial distortions, while spherical harmonic basis would not. Spherical harmonic basis has isotropy and rotation invariance, using Spherical Harmonic Transform (SHT) to process spherical data has natural advantages. To this end, Bonev et al. (2023) introduced Spherical Fourier Neural Operators (SFNOs) based on SHT for learning operators on spherical geometries, demonstrating stable autoregressive, while retaining physically plausible dynamics. However, they limit themselves to equivariance in the continuous limit, and the power of nonlinear fitting in the frequency domain still needs to be explored. Moreover, they ignored the spectral bias (Chattopadhyay & Hassanzadeh, 2023; John Xu et al., 2020; McCabe et al., 2023; Rahaman et al., 2019) introduced by data-driven models.

In this work, we introduce the Gated Residual Spectral Attention (GRSA) to effectively leverage nonlinear information in the frequency domain, and develop a general neural operator to mitigate the accumulation of errors for long-term iterations caused by distortions and spectral bias on the sphere. Specifically, we use the SHT to extract spatial features of different scales and extend the Multi-Head Self-Attention (MHSA) to spectral domain to explore the correlation among them. Then, inspired by the Laplacian matrix in graph theory and the MEGA (Ma et al., 2023) model, we design GRSA module to utilize the spectral information and correct outliers caused by the spurious correlations across different scales. Moreover, a general neural operator named Spherical Harmonic Neural Operator (SHNO) was introduced to improve the accuracy and stability of long-term iterative forecasts. Experiments for Shallow Water Equations (SWEs) solving and medium-range global weather forecasting demonstrate the effectiveness of the proposed methods.

The remainder of this paper is as follows. Section 2 introduces the proposed MHSA, the GRSA and the SHNO. Section 3 presents the employed datasets and the experimental designs. Section 4 describes some universal factors that cause instability for data-driven models iterative forecasts through spherical SWEs solving and medium-range global weather forecasting and evaluates the proposed methods. Finally, Section 5 concludes this work.

## 2 Methods

### 2.1 Spectral Multi-Head Self-Attention

Self-attention is the core component of the Transformers (Alexey et al., 2021; Vaswani et al., 2017), and multiple heads are usually used to enhance the performance. The general form of MHSA in Vision Transformers can be written as follows (Han et al., 2023; You et al., 2023):

$$Q^m = xW_Q^m, K^m = xW_K^m, V^m = xW_V^m,$$

$$O_i^m = \sum_{j=1}^{N} \frac{Sim\left(Q_i^m, K_j^m\right)}{\sum_{j=1}^{N} Sim\left(Q_i^m, K_j^m\right)} V_j^m,$$

where $i \in \{1, \ldots, N\}$ refers to the i-th row, $m \in \{1, \ldots, M\}$ refers to the m-th head, $Q^m, K^m, V^m \in \mathbb{R}^{N\times d}$ are the query, key, and value matrices obtained by linearly projecting of the input N tokens $x \in \mathbb{R}^{N\times C}$, $W_Q^m, W_K^m, W_V^m \in \mathbb{R}^{C\times d}$ are learnable projection matrices and $Sim(\cdot, \cdot)$ denotes the similarity function. And modern Transformers usually adopt Softmax function to measure the similarities, where $Sim(\mathrm{Q}, \mathrm{V}) = e^{\frac{QK^T}{\sqrt{d}}}$, superscript $T$ means transpose.

To investigate the performance of the MHSA in the spectral domain, we implement a spectral multi-head self-attention (SMHSA). Since the formula of each attention head is the same, we take a single attention head as an example and its calculation formula is given as follows:

$$Q_{CV} = zW_{Q_{CV}}, K_{CV} = zW_{K_{CV}}, V_{CV} = zW_{V_{CV}},$$

$$O_{CV} = CSoftmax\left(\frac{Q_{CV}K_{CV}^H}{\sqrt{d}}\right) V_{CV},$$

where $Q_{CV}, K_{CV}, V_{CV} \in \mathbb{C}^{N\times d}$ are the query, key, and value matrices respectively, obtained by linearly projecting of the input N tokens $z = x + iy \in \mathbb{C}^{N\times C}$ with $x, y \in \mathbb{R}^{N\times C}$, $W_{Q_{CV}}, W_{K_{CV}}, W_{V_{CV}} \in \mathbb{C}^{C\times d}$ are learnable projection matrices, $CSoftmax = Softmax(\mathcal{R}) + iSoftmax(\mathcal{J})$ with $\mathcal{R}$ represent the real part, $\mathcal{J}$ represent the imaginary part, and $K_{CV}^H \in \mathbb{C}^{d\times N}$ is conjugate transpose of $K_{CV}$.

### 2.2 Gated Residual Spectral Attention with Parametric Laplacian Matrix

Previous studies have shown, that transformer-based models may easily learn spurious correlation in the data (Enström et al., 2024), and display limited robustness when the pre-training dataset is relatively small (Ghosal & Li, 2024). To address the potential spectral bias arising from spurious correlations in the SMHSA, we present a parameterized Laplacian matrix to model the interplay among different scales rather than the correlation among them.

In graph theory, the Laplacian matrix can be considered as the discrete analog of the Laplacian operator in multivariable calculus. It also represents the degree of difference between a vertex and its nearby vertex values. Laplacian matrix is defined as $\boldsymbol{L} := \boldsymbol{D} - \boldsymbol{A}$, where $\boldsymbol{A} \in \mathbb{C}^{m \times m}$ is a weighted adjacency matrix of $\boldsymbol{X} \in \mathbb{C}^{m \times n}$ and $\boldsymbol{D} \in \mathbb{C}^{m \times m}$ is the degree matrix with $\boldsymbol{D}_{ii} = \sum_{j=1}^{m} \boldsymbol{A}_{ij}$ and $\boldsymbol{D}_{ij} = 0$ if $i \neq j$. The adjacency matrix $\boldsymbol{A}$ and Laplacian matrix $\boldsymbol{L}$ is unknown when $\boldsymbol{X}$ is incompletely sampled, so parameterization is required for machine learning applications (Li et al., 2021; Zhemin Li et al., 2023). And there are there properties to note: (1) The Laplacian matrix $\boldsymbol{L}$ is positive semi-definite and the sum of each row equals zero (Zhemin Li et al., 2023); (2) The natural spectral information is usually piecewise smooth, so $\boldsymbol{L}$ should be nearly smooth; (3) The adjacency matrix $\boldsymbol{A}$ of a directed graph is usually a Hermitian matrix.

In this paper, we refer to (Zhemin Li et al., 2023), learning the adjacency matrix through an MLP and the SoftMax function. The difference is that through the multiplication of the learned lower triangular matrix and its conjugate transpose, we assurance the adjacency matrix to a Hermitian matrix. Then the parameterized Laplacian matrix was calculated by the adjacency matrix, and used as the attention coefficient. Additionally, we implemented a moving average to the parameterized

Laplacian matrix to combine information from current and previous layers using an adaptive weight, inspired by the MEGA (Ma et al., 2023). Furthermore, the gating mechanism in the gated attention unit (GAU) (Hua et al., 2022) is equipped to control the flow of information (Lai et al., 2019). The structure of GRAS is shown in Figure 1(b), and the formula for each attention head is as follows:

$$\boldsymbol{Y}_\ell = \boldsymbol{X}_\ell W_{Y_{CV}} + b_{Y_{CV}},$$

$$\boldsymbol{X}'_\ell = Cat(\boldsymbol{X}_\ell, reg),$$

$$\boldsymbol{B}_\ell = SoftMax\big(g(\boldsymbol{X}'_\ell)\big),$$

$$\boldsymbol{A}_\ell = tril(\boldsymbol{B}_\ell)(tril(\boldsymbol{B}_\ell))^H,$$

$$\boldsymbol{L}_\ell = \sigma(\alpha_\ell)(\boldsymbol{A}_\ell \cdot \boldsymbol{1}_{N\times N} \odot \boldsymbol{I}_N - \boldsymbol{A}_\ell) + (1 - \sigma(\alpha_\ell))\boldsymbol{L}_{\ell-1},$$

$$\boldsymbol{Z}_\ell = (\boldsymbol{L}_\ell \cdot \varphi(\boldsymbol{X}'_\ell W_{V_{CV}} + b_{V_{CV}}),$$

$$\boldsymbol{Z}'_\ell = \boldsymbol{Z}_\ell \odot \varphi\big(\boldsymbol{X}'_\ell W_{Q_{CV}} + b_{Q_{CV}}\big),$$

$$\boldsymbol{O}_\ell = Drop\big(\boldsymbol{Z}'_\ell W_{P_{CV}} + b_{P_{CV}}\big) + \boldsymbol{Y}_\ell,$$

where $\boldsymbol{X}_\ell, \boldsymbol{O}_\ell \in \mathbb{C}^{N\times C}$ means the input and output, $\ell = 1 \dots L$ is the number of network layers, $W_{Y_{CV}}, W_{V_{CV}}, W_{P_{CV}}, W_{Q_{CV}}, b_{Y_{CV}}, b_{V_{CV}}, b_{P_{CV}}, b_{Q_{CV}} \in \mathbb{C}^{C\times d}$ are learnable projection matrices and bias, $reg$ means the registers (Darcet et al., 2024), $Cat(\cdot,\cdot)$ represents concatenate, $Drop(\cdot)$ represents remove the registers, $g(\cdot): \mathbb{C}^{N\times C} \mapsto \mathbb{C}^{\mathrm{N\times d}}$ is an MLP, which aims to capture self-similarity in $\boldsymbol{X}'_\ell$, $tril(\cdot)$ mask the matrix with the lower triangular matrix, $H$ refers to conjugate transpose, $\boldsymbol{1}_{N\times N}$ is an $N \times N$ matrix whose entries are all 1s, $\boldsymbol{I}_N$ is the $N \times N$ identity matrix, $\odot$ is Hadamard product, $\varphi$ is the smooth maximum unit (SMU) (Biswas et al., 2022), $\sigma$ is the sigmoid function, $\alpha_n$ is a learnable parameter, and the parameterized

Laplacian matrix $\boldsymbol{L}_n$ can measures the similarity between rows of $\boldsymbol{X}'_\ell$, $\boldsymbol{L}_{\ell-1}$ is the parameterized Laplacian matrix from the previous layer.

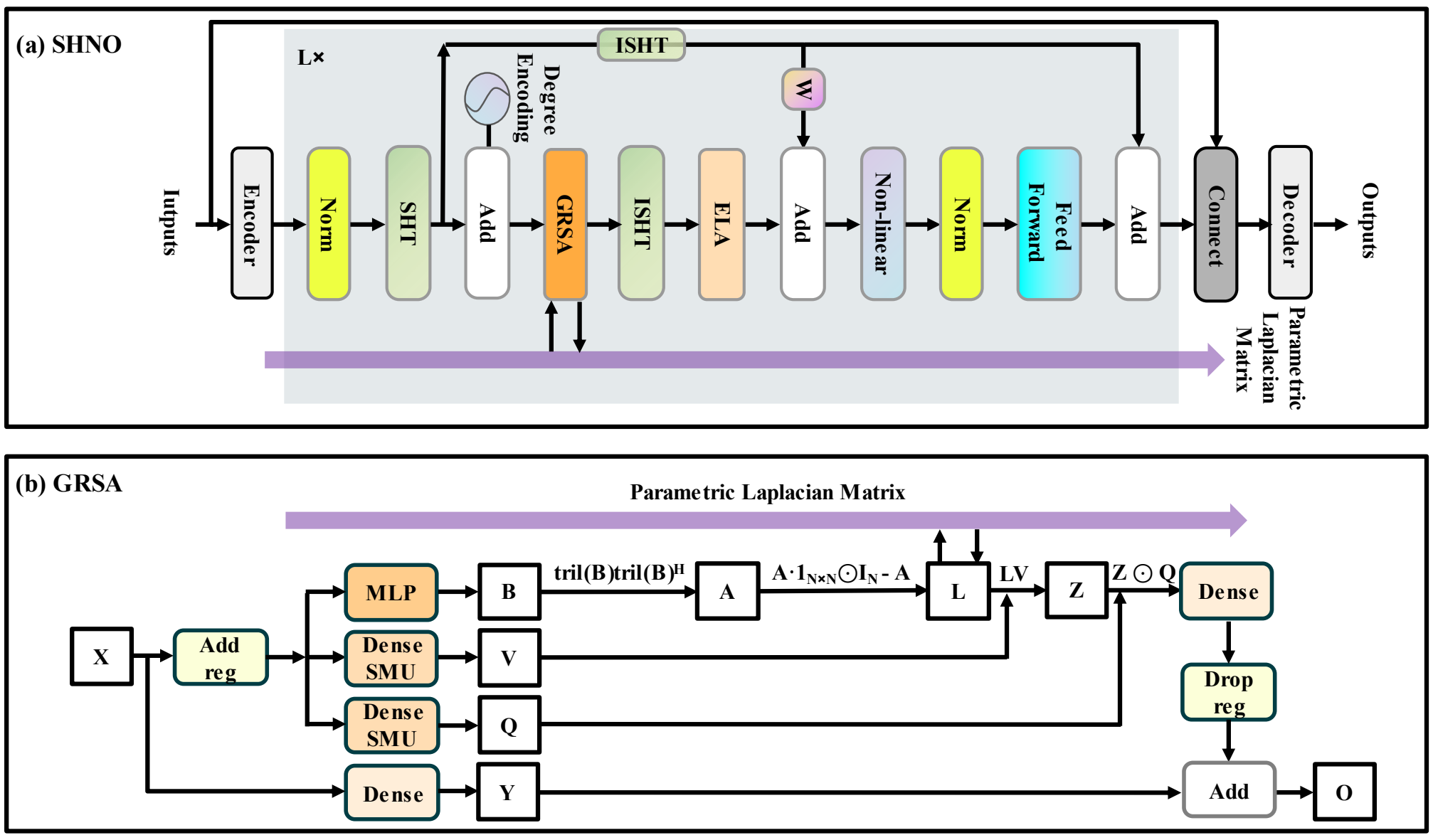

**Figure 1**: Overall architecture of the proposed model. (a) Spherical Harmonic Neural Operator (SHNO); (b) Gated Residual Spectral Attention (GRSA) with Parametric Laplacian Matrix.

### 2.3 Spherical Harmonic Neural Operator

To make the neural operators more stable and efficient for iterative forecasts, we construct Spherical Harmonic Neural Operator (SHNO) based on the SFNO (Bonev et al., 2023), ViT (Alexey et al., 2021) and the proposed GRSA. ViTs need to split the input data into several patches through a technique named patch embedding, since it can reduce the computational overhead and improve the adaptability. However, this common technique introduces discontinuity. In this paper, we remove the patch embedding, and use the total and the zonal wave number to reduce the computational complexity.

The architecture of SHNO is shown in the Figure 1(a). As we can see, the input is raised to the higher dimension channel space through the encoder, then the norm and the SHT is used to transform the data into spectral domain. This spectral coefficient matrix implies the degree information of the spherical harmonic functions, while the self-attention mechanism is insensitive to the degree information. To this problem, we add learnable degree encoding before the GRSA, inspired by the learnable position embeddings (Alexey et al., 2021). In the GRSA, we add registers (Darcet et al., 2024) to reduce the outliers of feature maps and improve the multi-step iteration performance. The operation in the spectral domain helps GRSA to capture global information easily. And we add the efficient local attention (ELA) (Xu & Wan, 2024) after GRAS to capture local information. Then the residual connection with a learnable parameter is used, according to (Ha & Lyu, 2022), this residual connection can help to correct the error caused by the numerical truncation of SHT and ISHT. After that, non-linear function, data norm and feed-forward network (FFN) is used. Conventional FFN integrates and maps global dependencies among different feature representations through a fully connected layer, which lacks local sensitivity (Shi et al., 2024). To this end, we choose the multi-path feed-forward network (MPFFN) (Shi et al., 2024) to integrate multi-scale dependencies. Finally, we use the decoder to transform the data back to the original dimension, and get the forecasts.

Let the input signal is $\boldsymbol{x} \in \mathbb{R}^{C \times H \times W}$, and the output is $\boldsymbol{y} \in \mathbb{R}^{C \times H \times W}$, then the formula of SHNO is:

$$\boldsymbol{z}_0 = Encoder(\boldsymbol{x}),$$

$$\boldsymbol{z}'_\ell = \mathcal{F}(Norm(\boldsymbol{z}_{\ell-1})),$$

$$\mathbf{z}''_\ell = ELA\left(\mathcal{F}^{-1}\left(GRSA(\mathbf{z}'_\ell + \mathbf{E}_{degree})\right)\right) + \mathcal{F}^{-1}(\mathbf{z}'_\ell)W,$$

$$\mathbf{z}_\ell = FFN\left(Norm\left(GELU(\mathbf{z}''_\ell)\right)\right) + \mathcal{F}^{-1}(\mathbf{z}'_\ell),$$

$$y = Decoder(Cat(\mathbf{z}_L, x))$$

where the $\mathbf{E}_{degree} \in \mathbb{R}^{C\times H\times W}$ is a learnable degree encoding, $\ell = 1 \dots L$ is the number of network layers, $\mathcal{F}$ and $\mathcal{F}^{-1}$ represents the SHT and ISHT, $Norm$ means the instance normalization (Ulyanov et al., 2016), $GELU$ is the GELU function, and $FFN$ is MPFFN.

# 3 Data and Experiments

## 3.1 Spherical Shallow Water Equations

The Shallow Water Equations (SWEs) on rotating sphere are a nonlinear hyperbolic PDEs system (Bonev et al., 2023), which are derived by integrating the Navier-Stokes equations over the depth of the fluid layer when the horizontal length scale is much larger than the vertical length scale. They are formulated as follows:

$$\begin{cases} \dfrac{\partial\varsigma}{\partial t} = -\dfrac{1}{a\cos\theta}\dfrac{\partial}{\partial\lambda}[(\varsigma+f)u] - \dfrac{1}{a\cos\theta}\dfrac{\partial}{\partial\theta}[(\varsigma+f)v\cos\theta], \\ \dfrac{\partial\delta}{\partial t} = \dfrac{1}{a\cos\theta}\dfrac{\partial}{\partial\lambda}[(\varsigma+f)v] - \dfrac{1}{a\cos\theta}\dfrac{\partial}{\partial\theta}[(\varsigma+f)u\cos\theta] - \nabla^2\left[\varphi + \dfrac{1}{2}(u^2+v^2)\right], \\ \dfrac{\partial\varphi}{\partial t} = -\dfrac{1}{a\cos\theta}\dfrac{\partial(\varphi u)}{\partial\lambda} - \dfrac{1}{a\cos\theta}\dfrac{\partial(\varphi v\cos\theta)}{\partial\theta} - \bar{\varphi}\delta. \end{cases}$$

where $f = 2\Omega\sin\theta$ is the Coriolis parameter with $\Omega$ being the angular velocity of the sphere, $\varsigma, \delta, \varphi, \bar{\varphi}, u, v, a$ are vorticity, divergence, geopotential height, mean geopotential height, the λ- and the θ-components of the velocity vector in the spherical coordinates, and the radius of the sphere, respectively. As a simplification of fluid motion equation, SWEs are widely used in atmospheric dynamics, tidal motion, tsunami propagation and the simulation of Rossby waves and Kelvin waves. The precision in addressing the SWEs serves as a crucial criterion for evaluating the efficacy and robustness of numerical solution methods.

We choose the parameters of the Earth as the parameters of the SWEs on rotating sphere, and the initial conditions of the geopotential height and velocity fields are generated by the Gaussian random fields. The parameters for initializing the geopotential height and velocity fields are set consistent with (Bonev et al., 2023). Specifically, the average value, the standard deviation of the initial layer depth, and the average value, the standard deviation of the initial velocity are $\varphi_{avg} = 10^3 g$, $\varphi_{std} = 120g$, and $0$, $0.2\varphi_{avg}$, respectively.

After setting the PDEs parameters and initial values, we use a classical spectral solver (Giraldo, 2001) to generate the numerical solutions with a spatial resolution of 256 × 512 and timesteps of 60 seconds. We use 128 initial conditions to simulate 240 hours and remove the first 100 hours of simulation because of the spin-up problem. Then the numerical solution is resampled to 64 × 128, and the training set is constructed with the remaining 140 hours. The solutions of the previous hour are used as the input of the model, and the solutions of the current moment are used as labels. The testing set uses 32 initial conditions, with 100 simulation hours burn-in.

We choose 20% of the training data as the validation data, and train the U-Net (Ronneberger et al., 2015), FourCastNet (Pathak et al., 2022), SFNO Linear (Bonev et al., 2023), SFNO Non-Linear (Bonev et al., 2023), SHNO-SMHSA and SHNO on the remaining training data for 50 epochs, respectively. The SHNO-SMHSA model was obtained by replacing the attention mechanism of the SFNO in the spectral domain with SMHSA. And the best weight was saved according to the validation data, their performance was compared on the testing data.

The batch size for training is 16, and the initial learning rate is 0.001, with a cosine decay reduced to 0.00002 at the end. The loss function is the mean geometric relative norm on the sphere of each channel, which formulate is (Bonev et al., 2023):

$$\mathcal{L}[F_{\vartheta}[u_n], u_{n+1}] = \frac{1}{3} \sum_{c \in \text{ channels}} \left( \frac{\sum_{i \in \text{ grid}} \; w_i |F_{\vartheta}[u_n](x_i) - u_{n+1}(x_i)|^2}{\sum_{i \in \text{ grid}} \; w_i |u_{n+1}(x_i)|^2} \right)^{\frac{1}{2}}$$

where $F_{\vartheta}[u_n]$ is the predicted by deep learning models and $u_{n+1}$ is the ground truth, $w_i$ are the products of the Jacobian $\sin\lambda_i$ and the quadrature weights.

### 3.2 Data and Experiment of Global Weather Forecast

The data we use for iterative medium-range global weather forecast is WeatherBench (Rasp et al., 2020), its publicly available at https://github.com/pangeo-data/WeatherBench. WeatherBench contains regirded ERA5 (Hersbach et al., 2020) data from 1979 to 2018 at hourly temporal resolution, and has three spatial resolutions to choose from: 5.625° (32×64 grid points), 2.8125° (64×128 grid points) and 1.40525° (128×256 grid points). Due to the limitation of our computing resources, 5.625° was chosen as the spatial resolution, and following previous studies (L. Chen et al., 2023; Lam et al., 2023), 6h was chosen as the minimum time resolution for the iterative forecast.

We use data from 1979 to 2015 as the training set, data from 2016 as the validation set, and data from out-of-sample, i.e., data from 2017 to 2018, as the testing set. And there are 22 variables for iterative forecasts, which are 10U, 10V, T2M, U1000, V1000, Z1000, U850, V850, Z850, T850, RH850, U500, V500, Z500, T500, RH500, U250, V250, Z250, T250, T100, Z50 respectively. The abbreviations and their descriptions are shown in supporting information (Table S3). Additionally, the input to the model contains two constant fields: the land-sea mask and the orography.

We use supervised training to predict a single time step on the training dataset. The loss function we choose is the latitude-weighted $\mathcal{L}2$ loss, which is defined as follows:

$$\mathcal{L}2[F_\vartheta[u_n], u_{n+1}] = \frac{1}{C \times H \times W} \sum_{c=1}^{C} \sum_{i=1}^{H} \sum_{j=1}^{W} w_i \left( F_\vartheta[u_n](x_{c,i,j}) - u_{n+1}(x_{c,i,j}) \right)^2$$

where $C, H, W$ are the number of channels, grid points in latitude, grid points in longitude, respectively. $F_\vartheta[u_n](x_{c,i,j})$ and $u_{n+1}(x_{c,i,j})$ are predicted and ground truth for same variable and latitude longitude coordinates at time step of $n+1$. $w_i$ is the latitude weighting factor for the latitude at the $i$th latitude index, which is calculated as follows:

$$w_i = \frac{\cos(\text{lat}(i))}{\frac{1}{H}\sum_i^H \cos(\text{lat}(i))}$$

where $\cos$ is the cosine function.

The models are developed using the PyTorch framework (Paszke et al., 2017) and utilize the training workflow provided by ClimaX (Nguyen et al., 2023). The models are trained with 100 epochs using a batch size of 80 on a single Nvidia GeForce RTX 3090 GPU. The initial learning rate is $2.0 \times 10^{-4}$, with a linear warmup schedule for 6 epochs, followed by a cosine-annealing schedule (Loshchilov & Hutter, 2017b) for 94 epochs. In addition, the AdamW (Kingma & Ba, 2014; Loshchilov & Hutter, 2017a) optimizer with parameters $\beta_1 = 0.9, \beta_2 = 0.99$ and weight decay of $1.0 \times 10^{-5}$, and bfloat16 floating point precision are applied for training.

The model with the lowest latitude-weighted RMSE on the validation set is saved, and evaluate on the test set. The evaluation metrics latitude-weighted RMSE and ACC are calculated as follows (Rasp et al., 2020; Rasp & Thuerey, 2021):

$$\mathrm{RMSE} = \frac{1}{N_{\mathrm{forecasts}}} \sum_{n}^{N_{\mathrm{forecasts}}} \sqrt{\frac{1}{N_{\mathrm{lat}} N_{\mathrm{lon}}} \sum_{i}^{N_{\mathrm{lat}}} \sum_{j}^{N_{\mathrm{lon}}} w_i \left(f_{n,i,j} - t_{n,i,j}\right)^2}$$

$$\mathrm{ACC} = \frac{\sum_{n,i,j} w_i f'_{n,i,j} t'_{n,i,j}}{\sqrt{\sum_{n,i,j} w_i f'^2_{n,i,j} \sum_{n,i,j} w_i t'^2_{n,i,j}}}$$

where $f$ is the model forecast and $t$ is the ERA5 truth, $w_i$ is the latitude weighting factor for the latitude at the $i$th latitude index, the prime ' denotes the difference to the climatology and the climatology is defined as $\mathrm{climatology}_{i,j} = \frac{1}{N_{\mathrm{time}}} \sum t_{i,j}$.

## 4 Results

This section will first describe some universal factors that cause instability for data-driven models solving spherical SWEs. Then, we will present a method to overcome these challenges and enhance the accuracy of long-term predictions. Finally, the main findings and the proposed method will verified by ERA5 data and popular data-driven medium-range global weather forecasting.

### 4.1 Spherical Shallow Water Equations

We begin by demonstrating experimentally that conventional convolution models cause distortions when processing spherical data. The widely used and high-performing U-Net (Ronneberger et al., 2015) and FourCastNet (Pathak et al., 2022) were selected as the baseline of conventional convolution, SFNO Linear (Bonev et al., 2023), and SFNO Non-Linear (Bonev et al., 2023) as the baseline of spherical convolution. As shown in Figure 2(a)-(d), the relative errors of geopotential height forecasted by U-Net are obvious near the poles and on the east-west boundary, even at the initial iteration. With the number of iterations increases, the errors propagate from the poles to the mid and low latitudes and gradually encompass the entire domain. This phenomenon is caused by distortions at the poles and the zero-padding at the boundaries seriously affects the accuracy. FourCastNet uses Discrete Fourier Transform (DFT) to ensure the continuity of the east-west boundary and does not need paddings, obtaining minimum relative errors for one-step prediction. However, due to the implicit flat hypothesis of DFT and the period in the meridian direction, errors initially concentrated near the poles rapidly spread to the entire domain with an increase in iteration steps (Figure 2 and Figure S1 in supporting information). To overcome this limitation, SFNO (Bonev et al., 2023) introduced SHT to the data-driven models. As shown in Figure 2(i)-(p), using SHT can ensure the continuity of the east-west boundary and reduce the distortions at the poles. Although the transformation from the Gaussian grid to the latitude and longitude grid still causes bias in the poles, it is well-limited and has small effects on other domains. And in the first iteration, the errors of SFNO Linear are smaller than those of SFNO Non-Linear. But as the number of iterations increases, SFNO Non-Linear gradually outperforms SFNO Linear. The key difference between SFNO Non-Linear and SFNO Linear is whether the non-linearity functions were used in the frequency domain. SFNO Linear attempts to maintain trivially equivariant using point-wise operations under the assumption of continuous. However, in the discrete data, this method only keeps approximately equivariant (Bonev et al., 2023), and the effect of this error is intensified with the increase of iteration steps. SFNO Non-Linear, which applies non-linearities in the frequency domain, remedied this situation (Bonev et al., 2023; Poulenard & Guibas, 2021).

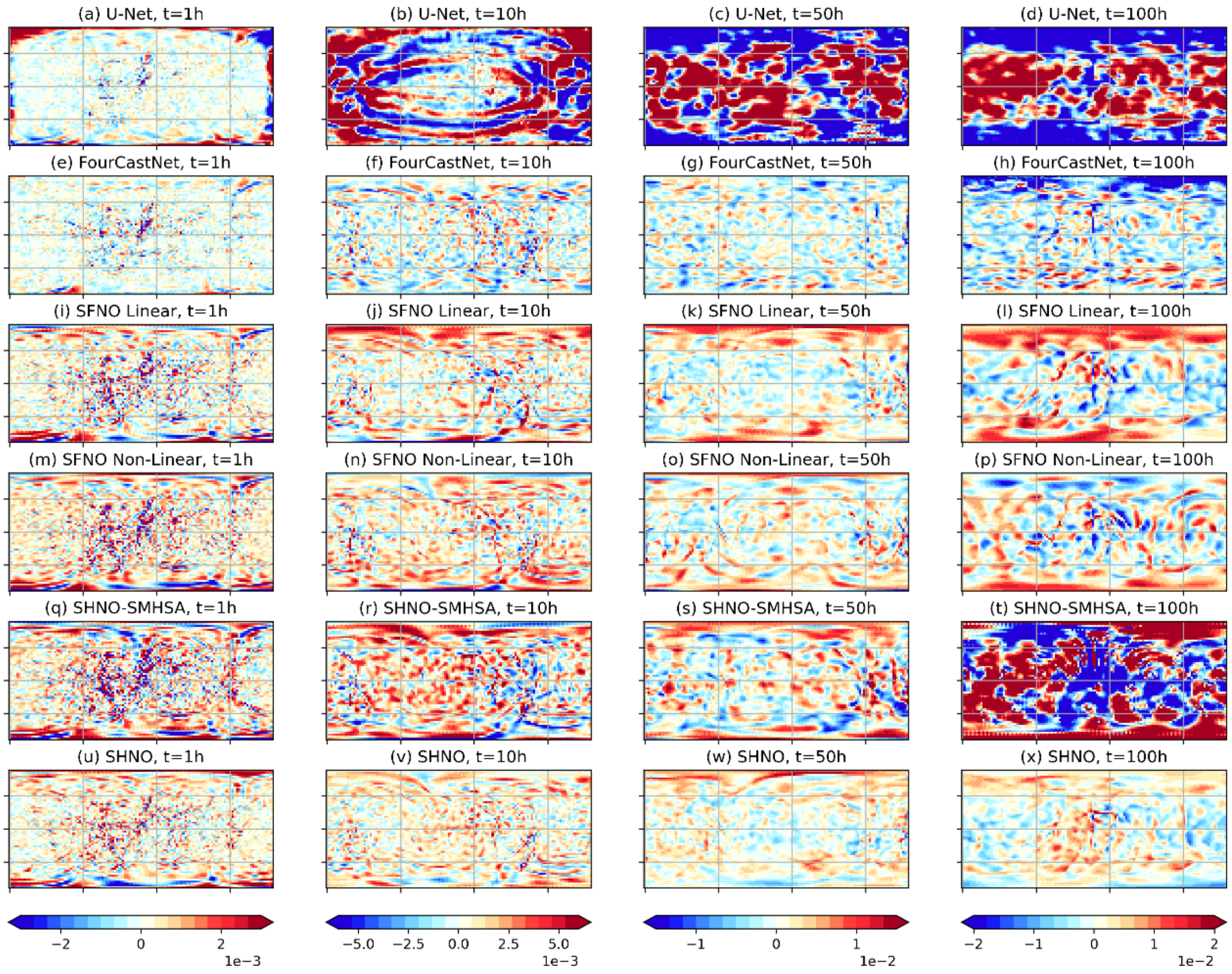

**Figure 2**: Spatial distribution of relative errors for geopotential height in Spherical

Shallow Water Equations. The smaller the absolute value, the better the performance. Columns from left to right corresponding to 1, 10, 50 and 100 iteration steps respectively. Rows from top to bottom represent the ground truth, U-Net, FourCastNet, SFNO Linear, SFNO Non-Linear, SHNO-SMHSA, and SHNO respectively.

Another factor that leads to the instabilities of iterative prediction is spectral bias (Chattopadhyay & Hassanzadeh, 2023; John Xu et al., 2020; McCabe et al., 2023; Rahaman et al., 2019). We use SHT to geopotential height $Z$ and calculate its spectra on different iteration steps, the calculation formula is as follows (Koshyk & Hamilton, 2001; Zongheng Li et al., 2023; Niranjan Kumar et al., 2023):

$$Z(\lambda, \varphi, p, t) = \sum_{n=0}^{N} \sum_{m=-n}^{n} Z_n^m(p,t) P_n^m(cos\varphi) e^{im\lambda},$$

$$E_n^m(Z,p,t) = \frac{1}{2} |Z_n^m(p,t)|^2$$

where $\lambda$, $\varphi$, $m$ and $n$ are longitude, latitude, zonal wavenumber, and total wavenumber, respectively. $N$ denotes truncated wavenumber, $Z_n^m$ is spectral coefficients of $Z$, $P_n^m$ represents associated Legendre polynomials with order $m$ and degrees $n$, $E_n^m(Z,p,t)$ is the geopotential spectra. As shown in Figure 3, in the first iteration, the spectra predicted by U-Net closely match the numerical labels, even at high frequency, which are typically considered challenging for data-driven models. After ten iterations, the low-frequency energy is underestimated, while the high-frequency energy is overestimated. With the increment of iteration steps, distortions induced by conventional convolution becomes more pronounced, leading to significant deviations in the spectra. When the iteration steps are small, FourCastNet also has a small spectral bias and good high-frequency fitting ability. However, when the iteration steps are large, data distortion causes the high-frequency energy to be falsely high and gradually affects the entire frequency. Although SHT can mitigate distortions and make the long-term iterative prediction more stable, the global nature of the spherical harmonic inevitably reduces the accuracy of localized features, which is shown as an underestimate of high-frequency energy in Figure 3. Furthermore, using non-linearities in the frequency domain is beneficial for stability.

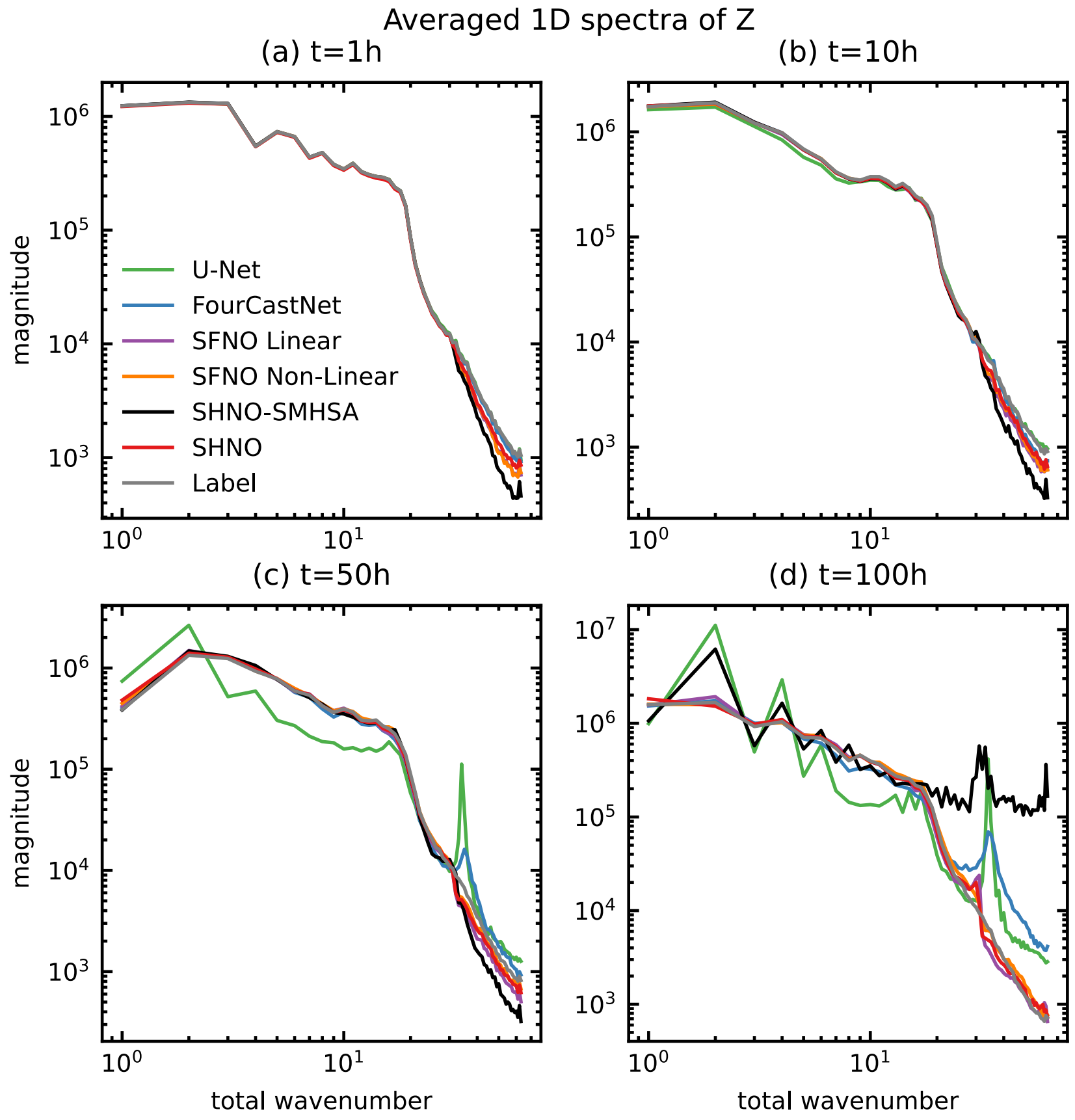

**Figure 3**: The geopotential spectra for different iteration steps (i.e. lead times) in Spherical Shallow Water Equations. The sign t in the figure represents the lead time: (a) 1 iteration step i.e. the lead time is 1h; (b) 10 iteration steps i.e. the lead time is10h; (c) 50 iteration steps i.e. the lead time is 50h; (d) 100 iteration steps i.e. the lead time is 100h.

Effectively simulating large-scale and mesoscale dynamics, and smoothing small-scale information seems to play a significant role in keeping long-term iterative stability. However, as shown in Figure 3, the excessive loss of small-scale information in SHNO-SMHSA leads to a deterioration in forecasting large-scale and mesoscale as the number of iterations increases. This, in turn, has a detrimental impact on the overall stability of the iterations. The SHNO-SMHSA is developed by replacing the attention mechanism of the SFNO in the spectral space with SMHSA. Simply extending MHSA into the spectral domain fails to enhance performance because the absence of small-scale information leads to an escalation in forecast noise and exacerbates the Gibb phenomenon (Figure 2). This not only impacts the accuracy of short-term iteration but also the stability of long-term iteration.

Therefore, it is imperative to mitigate distortions and spectral bias for long-term stable iterative forecasts. The SHNO, as introduced in this paper, uses SHT to alleviate the impact of distortions and employs efficient local attention (ELA) (Xu & Wan, 2024) and multi-path feed-forward network (MPFFN) (Shi et al., 2024) to rectify the underestimated small-scale information. Furthermore, to avoid the potential spectral bias arising from spurious correlations across different scales, we present a parameterized Laplacian matrix to model the interactions among these scales rather than the correlation among them. Additionally, SHNO also uses registers (Darcet et al., 2024) in spectral domain to eliminate outliers.

Table 1 shows the parameters of SHNO and the baselines. To demonstrate the efficacy of the proposed model, the embedding dimension used in SFNO Linear is twice that of SHNO. When the embedding dimension and model layers are the same, SHNO-SMHSA has the largest model parameters, followed by SFNO Non-Linear, while FourCastNet has the lowest parameters. The SMHSA and nonlinear modules in SFNO significantly increase the number of parameters.

**Table 1** Model parameters for spherical Shallow Water Equations.

| Model | Parameters | | |
|---|---|---|---|
| | Layers | Embed. dimension | Parameter count |
| U-Net | 5 | -- | 34.527M |
| FourCastNet | 4 | 512 | 8.941M |
| SFNO, Linear | 4 | 512 | 10.117M |
| SFNO, Non-Linear | 4 | 256 | 137.768M |
| SHNO-SMHSA | 4 | 256 | 139.873M |
| SHNO | 4 | 256 | 7.0572M |

Figure 4 shows the mean relative losses for the SWEs on the rotating sphere at a spatial resolution of 64 × 128 and a temporal resolution of 1 hour. As depicted, the mean relative loss of U-Net iterative forecasts accumulates rapidly, while FourCastNet demonstrates the most accurate initial forecasts (see Table S1 in supporting information for details). The initial forecasts of SFNO Linear are slightly better than SFNO Non-Linear, but the multi-step iterative forecasts are worse than it. These two models perform similarly, despite a significant difference in parameters. As SHNO-SMHSA shows, simply extending self-attention into the spectral domain fails to enhance performance, and the error explodes when the iteration steps are large. Nevertheless, SHNO-SMHSA outperforms U-Net when the iteration steps are less than 50. The SHNO performs best for long-term iterative forecasts, as it alleviates the impact of distortions and forecast noise (Figure 2) and enhances the precision and stability of small-scale simulations (Figure 3). Additionally, a comprehensive analysis of Figures 3 and 4 shows that the mean relative losses are primarily influenced by the mid and low frequency of the spectra, and the high frequency predominantly impacts the stability of iterative forecasts.

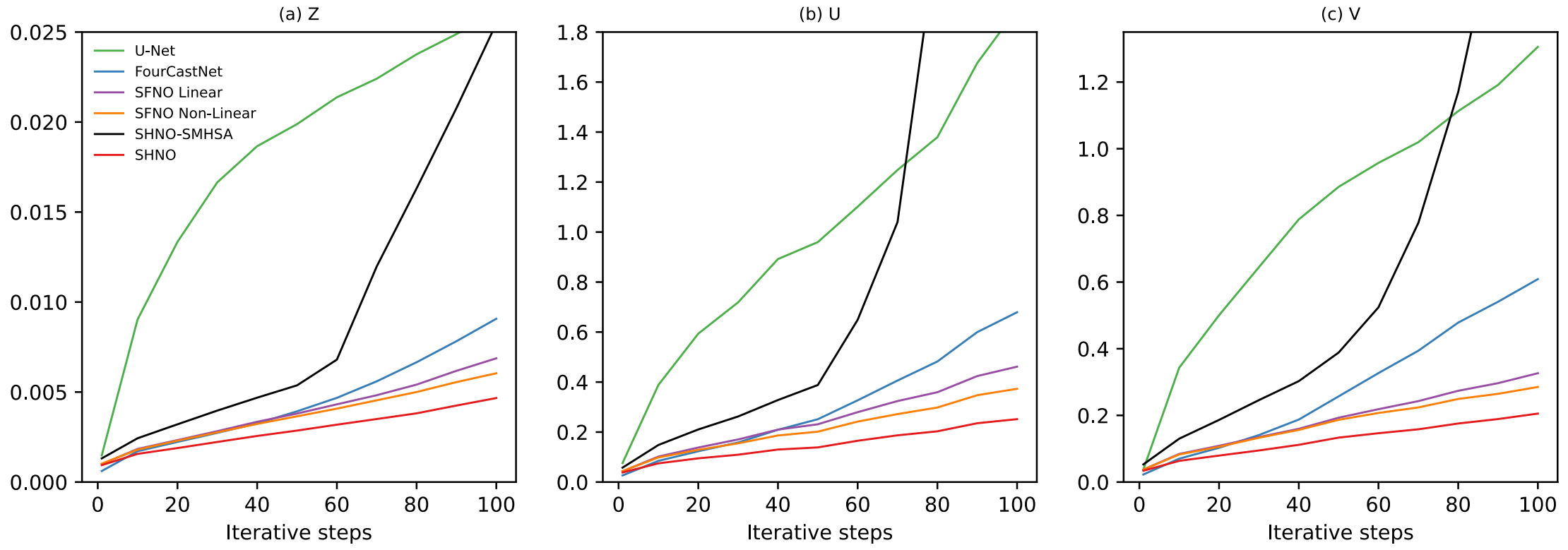

**Figure 4**: Mean relative losses for the Shallow Water Equations on the rotating sphere at a spatial resolution of 64 × 128 and a temporal resolution of 1 hour. Lower is better. The reported iteration steps are from 1 to 100 i.e. the lead time are from 1h to 100h. (a) mean relative loss of geopotential height $Z$; (b) mean relative loss of zonal wind velocity $U$; (c) mean relative loss of meridional wind velocity $V$.

### 4.2 Medium-Range Global Weather Forecast

To explain the potential effects of data distortions and spectral bias for data-driven weather forecast models, this study uses widely noticed FourCastNet (Pathak et al., 2022), Pangu-Weather (Bi et al., 2023), and FuXi (L. Chen et al., 2023) models for iterative forecasts. Figure 5 shows the spatial distribution of forecasts relative errors for geopotential height at 500hPa pressure level with a temporal resolution of 6 hours and a spatial resolution of 0.25° × 0.25°, and the input time is 00:00 UTC on 1 September 2018. When using FourCastNet for iterative forecast, Nan (not a number) values appear soon near the South Pole, indicating that there have been severe distortions. Although Nan values do not appear in the other two models, the maximum relative errors are also concentrated near the poles. And similarly to the results of spherical SWEs, the distortions also intensify as the number of iterations increases and gradually affect the mid and low-latitude domains. Furthermore, because we did not use the hierarchical temporal aggregation strategy (Bi et al., 2023), the relative errors of Pangu-Weather accumulated rapidly with the number of iterations. This phenomenon is not only associated with the architecture of the model but also with the forecasting strategies employed. Pangu-Weather can get accurate forecasts of specific lead times but ignores the stability of iterations. Therefore, it needs the hierarchical temporal aggregation strategy to reduce the iteration steps and control the accumulation of errors in long-term forecasts. Although the hierarchical temporal aggregation strategy leads to considerable performance gains, it suffers from temporal inconsistency (L. Chen et al., 2023). To settle this problem, FuXi presents a cascade model that balances accuracy for specific lead times with the stability of iterations. However, a single model of FuXi only guarantees stability for specific 5-day forecast times. When utilizing FuXi-Short for 7-day forecasts, as Figure 5(l) shows, the error accumulates fast. In general, none of them addresses the effects of spherical data distortion.

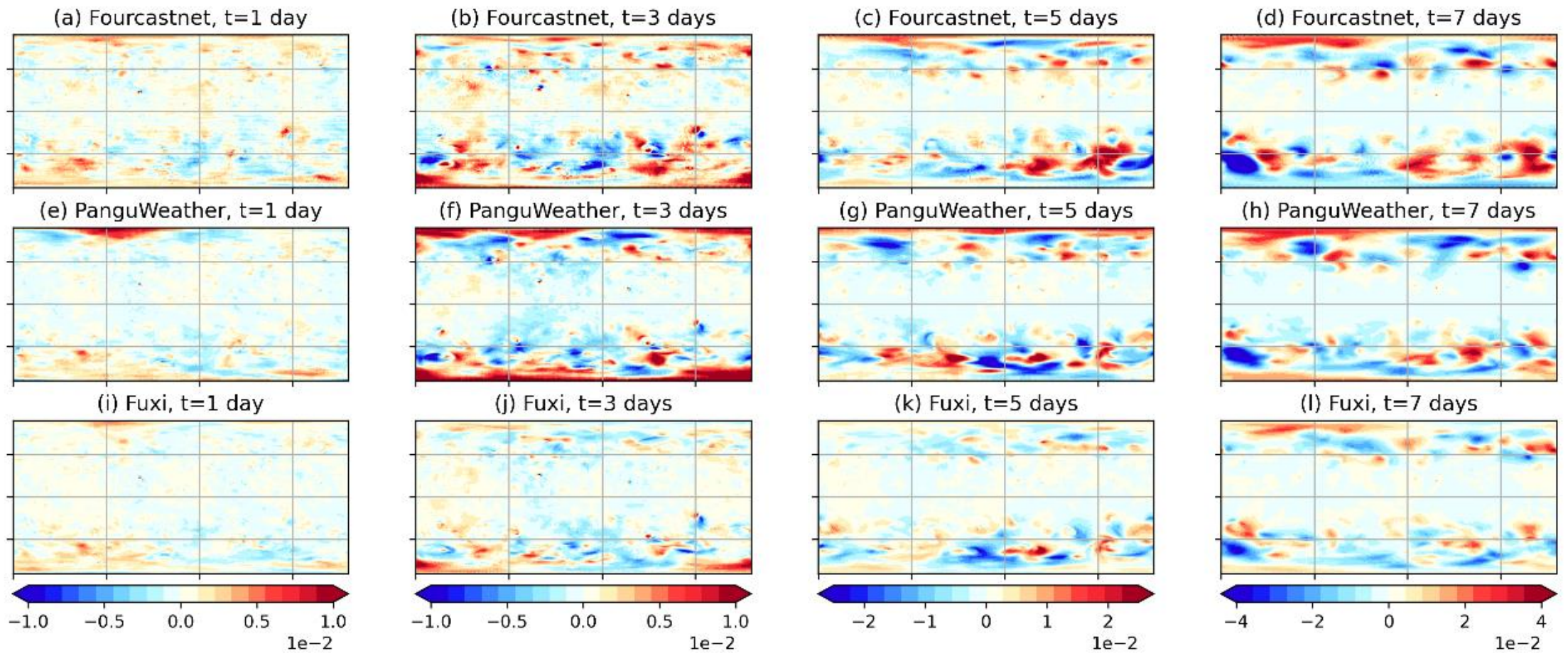

**Figure 5**: Spatial distribution of relative errors for geopotential height at 500hPa

pressure level. Columns from left to right correspond to 1 day, 3 days, 5 days, and 7 days lead time i.e. 4, 12, 20, and 28 iteration steps respectively. Rows from top to bottom represent the FourCastNet, Pangu-Weather, and FuXi models. For all cases, the input time is 00:00 UTC on 1 September 2018, and the spatial resolution is 0.25° × 0.25°.

Figure 6 shows the geopotential spectra of different models in Medium-Range Global Weather Forecast with spatial resolution of 0.25° × 0.25°. It has been noted that the distortions around the poles lead to an overestimation of energy at small-scales in iterative forecasts. Over time, these biases will impact both the mesoscale and large-scale, leading to a decline in the accuracy of long-term predictions. Although the spectra predicted by the FuXi model exhibit remarkable proximity with ERA5, spurious spikes become apparent when the total wave number is around 180, 360, 540, and 720. These sudden deviations are more obvious in the upper atmosphere, while their locations remain unchanged with iterations. This phenomenon indicates that there are systematic errors in the FuXi model. We also found that FourCastNet has similar deviations, and these discontinuities occurred in the same location as Fuxi. Similarly, the spectra predicted by Pangu-Weather also show significant fluctuation at the same location. All these models used a common technique named patch embedding for dimensionality reduction, which does introduce discontinuity.

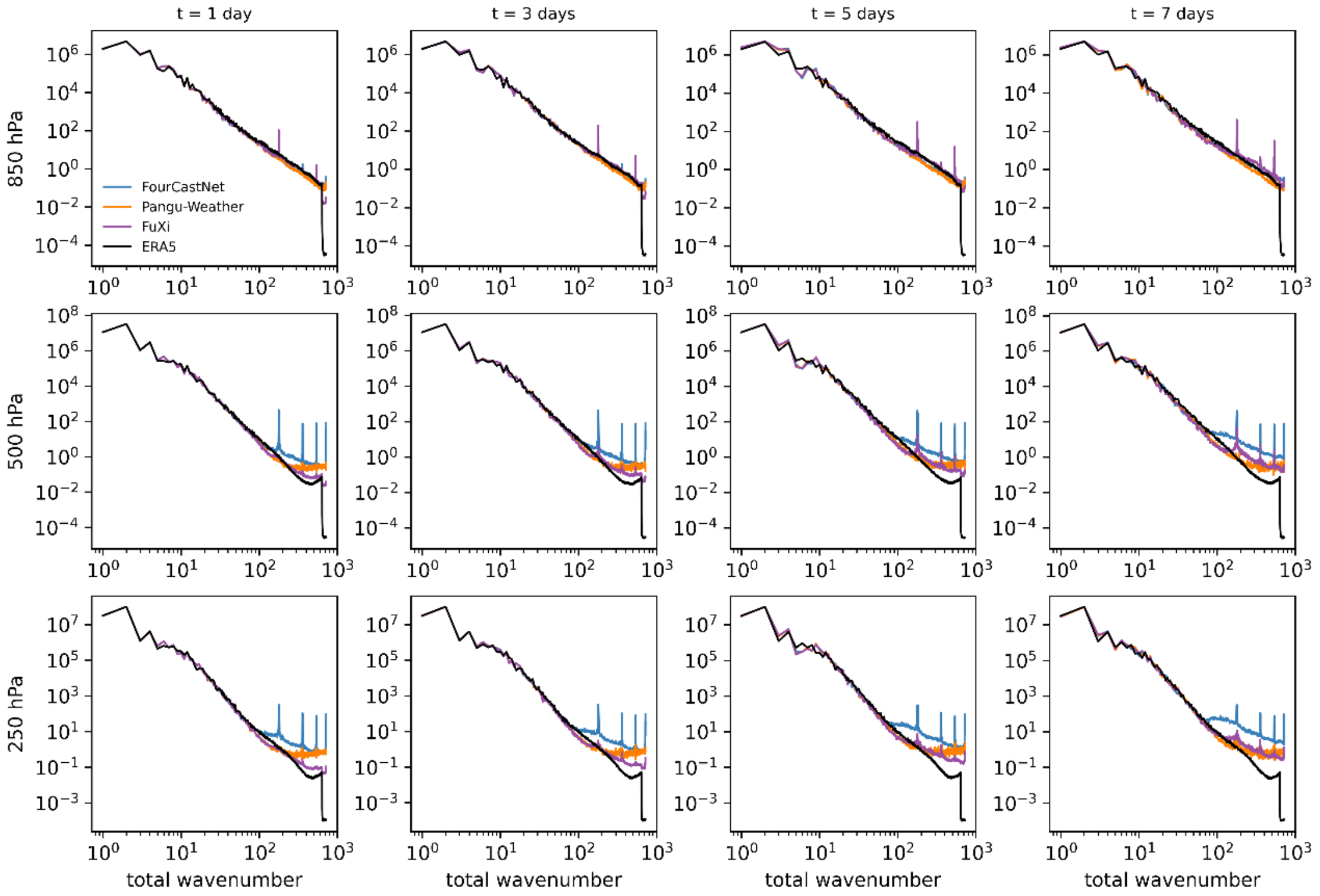

**Figure 6**: The geopotential spectra for different models with spatial resolution of 0.25° × 0.25°. Columns from left to right correspond to 1 day, 3 days, 5 days, and 7 days lead time i.e. 4, 12, 20, and 28 iteration steps respectively. Rows from top to bottom represent the 850hPa, 500hPa, and 250hPa pressure level. For all cases, the input time is 00:00 UTC on 1 September 2018.

To sum up, the advanced data-driven weather forecast models still suffer spectra bias and distortions near the poles. Furthermore, the essential technique named patch embedding in vision transformers also leads to discontinuity in forecasts. So, we use SHT to correct the distortions and remove the patch embedding, while using the total and the zonal wave number to control the computational complexity. To prove the validity of the proposed method, we use WeatherBench and the same training parameters to train FourCastNet, SFNO (Linear), and the proposed SHNO model. The parameters of the models are around 21M (see Table S2 in supporting information for details), and their performance was compared with IFS T42 on the test set.

Figure 7 shows the globally-averaged latitude-weighted RMSE of different models for different lead times of 3 surface variables (T2M, U10, and V10), 4 upper-air variables (Z500, T500, U500, and V500) at 500hPa pressure level, and 5 upper-air variables (Z850, T850, U850, V850, and RH850) at 850hPa pressure level. When the lead time is small, the performance of FourCastNet and the other models is close. However, the RMSE of FourCastNet increases rapidly with the number of iterations increases. For example, the geopotential at 850hpa pressure level predicted by FourCastNet initially outperforms the IFS but fell behind when the lead time exceeded three days (similar to (Pathak et al., 2022)). DFT in FourCastNet leads to distortions near the poles, which further affect the accuracy and stability of iterative forecasts. Therefore, FourCastNet need to tune the parameters or improve the iterative strategy. The RMSE of SHNO is smaller than SFNO and outperforms the IFS at the surface and 850hPa pressure level. Although the RMSE of wind predicted by SHNO is worse than IFS at 500hPa pressure level, it tends to catch up as the lead time extends. Furthermore, Figure S11 shows the globally-averaged latitude-weighted ACC, which also demonstrates that SHNO has better long-term iterative ability than FourCastNet and SFNO.

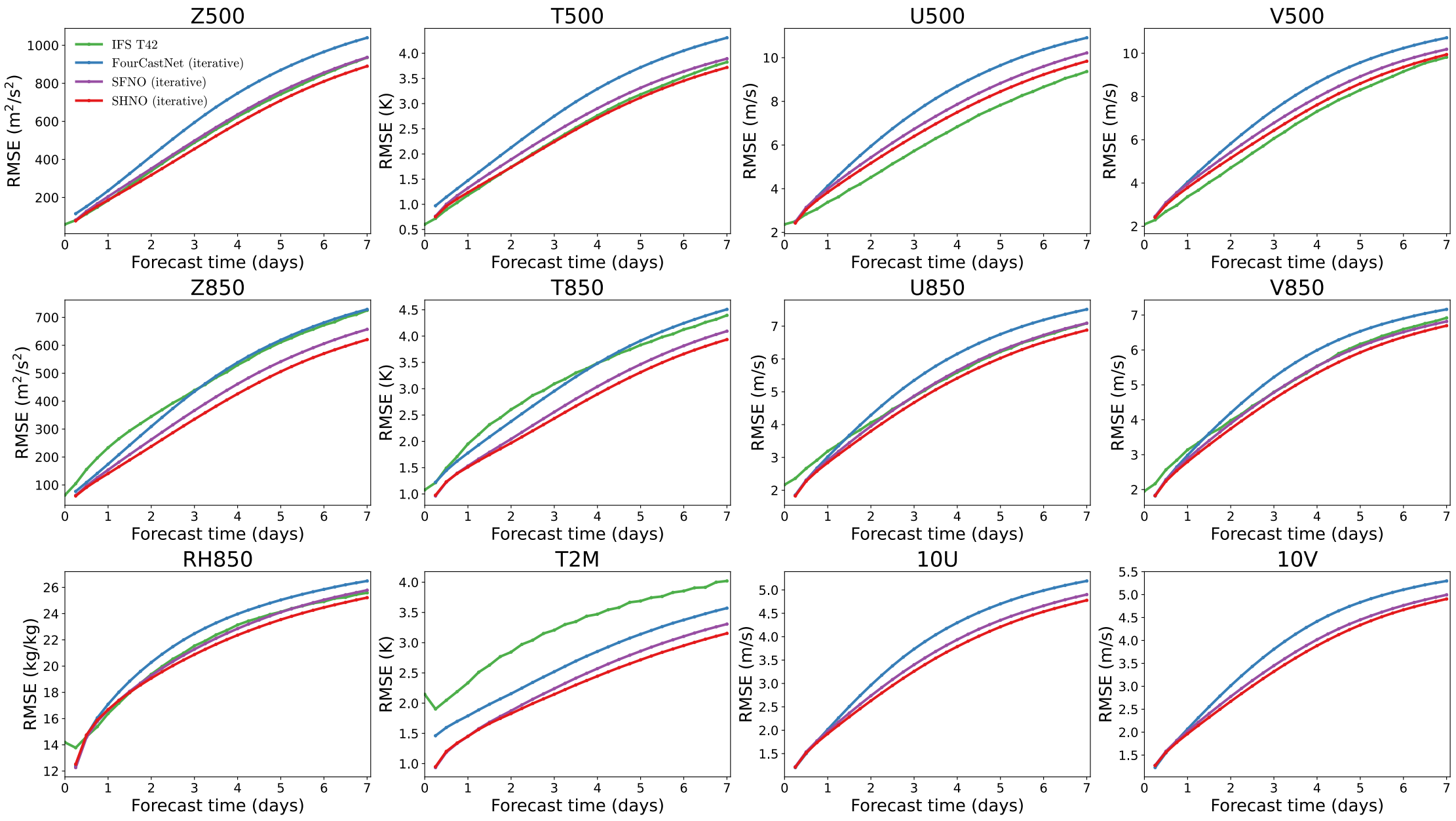

**Figure 7**: Globally averaged latitude-weighted RMSE of IFS T42 (green line), FourCastNet (blue lines), SFNO (purple lines), and SHNO (red lines) for 3 surface variables, 4 upper-air variables at 500hPa pressure level, and 5 upper-air variables at 850hPa pressure level with spatial resolution of 5.625° × 5.625° in 7 days forecasts using testing data from 2017 to 2018. Lower is better. IFS T42 were not available for 10U and 10V.

The spectra of geopotential for 1 day, 3 days, 5 days, and 7 days at the 500hPa pressure level, 850hPa pressure level, and 1000hPa pressure level are shown in Figure 8. The spectra at 1000hPa of IFS display significant deviation in mid and low frequency whereas the 500hPa is close to the actual situation. As lead time increases, the deviation of SFNO spreads gradually from high to mid and low frequency, while SHNO still show a good fit in mid and low frequency. Although patch embedding enables FourCastNet to retain more small-scale information, the effect of distortion leads to an overestimate of the small-scale energy for geopotential and destroys the stability of long-term iterations. The models using spherical harmonic transformation smooth the small-scale information and obtain more stable iterative forecasts.

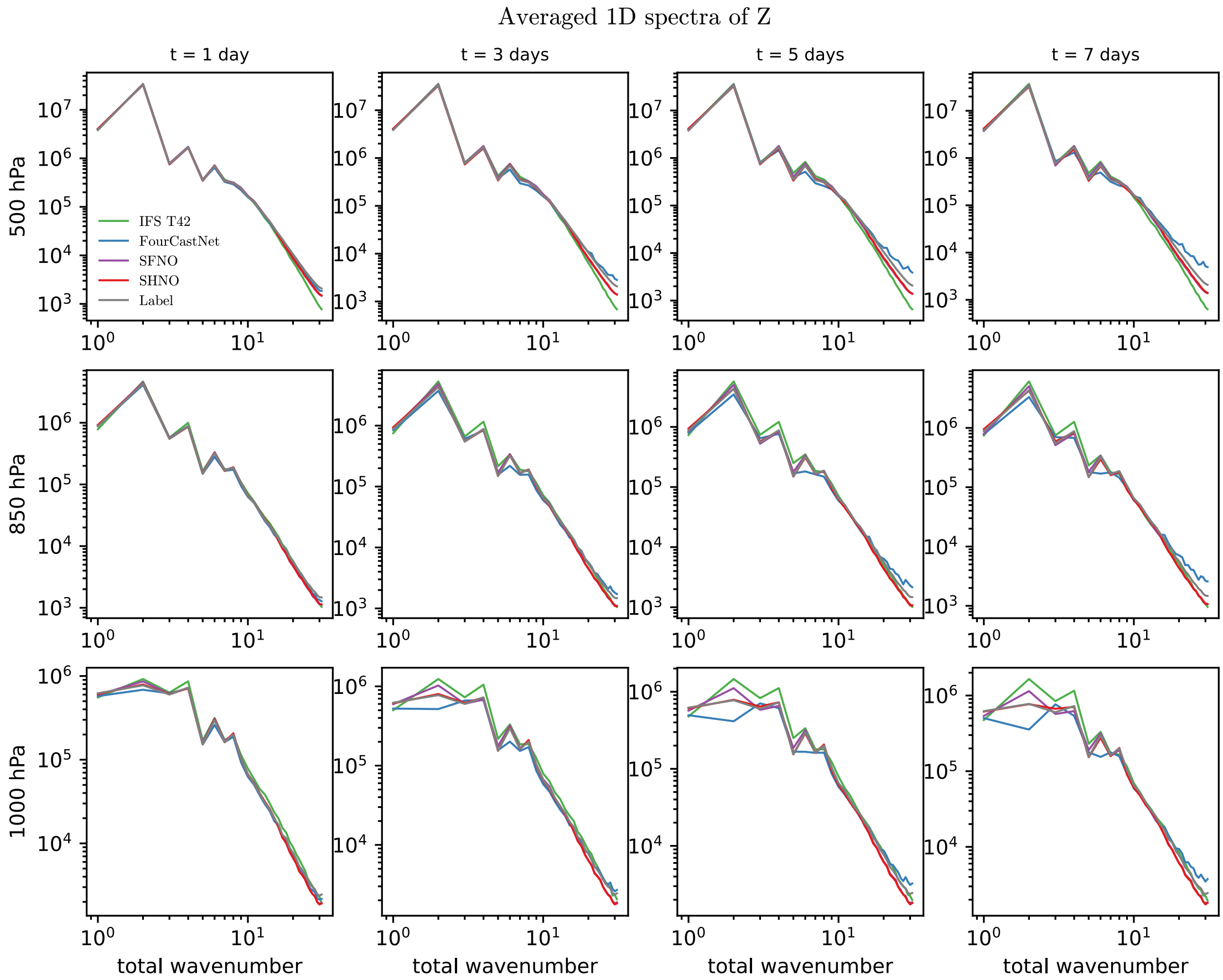

**Figure 8:** The geopotential spectra for different models with spatial resolution of 5.625° × 5.625°. The sign t in the figure represents the lead time. Columns from left to right corresponding to 4, 12, 20 and 35 iteration steps, respectively. Rows from top to bottom represent the 500hPa pressure level, 850hPa pressure level and 1000hPa pressure level.

## 5 Conclusion

Recently, data-driven weather forecasting systems (Bi et al., 2023; K. Chen et al., 2023; L. Chen et al., 2023; Lam et al., 2023) have shown stronger deterministic forecast results than the ECMWF's IFS (Bougeault et al., 2010). Nevertheless, information loss in small-scale of conventional neural networks causes spectral bias, leading to long-term iteration instability. We found that the universal mechanism for these instabilities is not only related to spectral bias but also to distortions brought by processing spherical data using conventional convolution. These distortions lead to a rapid amplification of errors over successive long-term iterations, resulting in a significant decline in forecast accuracy. The severe distortions first appear near the poles, and gradually propagate to the mid and low latitudes as the number of iterations increases. Furthermore, the discontinuity arising from patch embedding in transformer-based weather forecast models (such as FourCastNet (Pathak et al., 2022), Pangu-Weather (Bi et al., 2023), and FuXi (L. Chen et al., 2023)) increases the spectral bias. The combined effects of distortions and spectral bias severely affect the accuracy and stability of the long-term iteration. Aggregate multiple models can achieve optimal performance across various lead times, but they still suffer artificial distortions and may introduce temporal inconsistency (L. Chen et al., 2023).

To address the above challenges, we present a universal neural operator named SHNO to mitigate these distortions and spectral bias. SHNO uses the spherical harmonic basis (same as SFNO (Bonev et al., 2023)) to mitigate distortions for spherical data and uses GRSA with parametric Laplacian matrix to explore the interaction between different scales, which reduces the spectral bias caused by spurious correlation. Furthermore, we also use ELA (Xu & Wan, 2024) and MPFFN (Shi et al., 2024) to alleviate the loss of small-scale information. Experiments for spherical SWEs solving and medium-range global weather forecasting demonstrate that SHNO improves the accuracy and stability for long-term iterative forecasts. Furthermore, we found that directly extending the vanilla multi-head self-attention to the spectra domain fails to improve the performance while losing more small-scale information. Although proper smoothing of small-scale information is believed to contribute to long-term iterative stability, our experiments show that excessive loss of small-scale information worsens the spectral bias as the number of iterations increases. Therefore, small-scale information simulation and iteration stability need to be balanced.

Despite the promising accuracy and stability for long-term iteration, SHNO still has some limitations. First, the model is biased to generate smooth forecast results and tends to underestimate energy on small-scales. Second, for wind forecasts at 500hPa pressure level, the SHNO is inferior to the ECMWF's IFS, due to the limited model parameters and few training variables.

In future work, we will increase the network's depth and width, incorporate a more extensive range of variables, and employ higher-resolution data to train our model. In addition, we will find a good balance between modeling high frequency information and creating stable long-term iterative forecasts. And the new experimental results will be compared with the state-of-art weather forecast models, such as Pangu-Weather (Bi et al., 2023), GraphCast (Lam et al., 2023) and FuXi (L. Chen et al., 2023) model.

## References

Alexey, D., Lucas, B., Alexander, K., Dirk, W., Xiaohua, Z., Thomas, U., Mostafa, D., Matthias, M., Georg, H., Sylvain, G., Jakob, U., & Neil, H. (2021). *An Image is Worth 16x16 Words: Transformers for Image Recognition at Scale* International Conference on Learning Representations, https://openreview.net/forum?id=YicbFdNTTy

Bauer, P., Thorpe, A., & Brunet, G. (2015). The quiet revolution of numerical weather prediction. *Nature*, *525*(7567), 47-55. https://doi.org/10.1038/nature14956

Bi, K., Xie, L., Zhang, H., Chen, X., Gu, X., & Tian, Q. (2023). Accurate medium-range global weather forecasting with 3D neural networks. *Nature*, *619*(7970), 533-538. https://doi.org/10.1038/s41586-023-06185-3

Biswas, K., Kumar, S., Banerjee, S., & Pandey, A. K. (2022, 18-24 June 2022). Smooth Maximum Unit: Smooth Activation Function for Deep Networks using

Smoothing Maximum Technique. 2022 IEEE/CVF Conference on Computer Vision and Pattern Recognition (CVPR),

Bonev, B., Kurth, T., Hundt, C., Pathak, J., Baust, M., Kashinath, K., & Anandkumar, A. (2023). Spherical Fourier Neural Operators: Learning Stable Dynamics on the Sphere. *Proceedings of the 40th International Conference on Machine Learning*, *202*, 2806--2823. https://proceedings.mlr.press/v202/bonev23a.html

Bougeault, P., Toth, Z., Bishop, C., Brown, B., Burridge, D., Chen, D. H., Ebert, B., Fuentes, M., Hamill, T. M., Mylne, K., Nicolau, J., Paccagnella, T., Park, Y.-Y., Parsons, D., Raoult, B., Schuster, D., Dias, P. S., Swinbank, R., Takeuchi, Y., Tennant, W., Wilson, L., & Worley, S. (2010). The THORPEX Interactive Grand Global Ensemble. *Bulletin of the American Meteorological Society*, *91*(8), 1059-1072. https://doi.org/https://doi.org/10.1175/2010BAMS2853.1

Chattopadhyay, A., & Hassanzadeh, P. (2023). Long-term instabilities of deep learning-based digital twins of the climate system: The cause and a solution. *arXiv e-prints*, *abs/2304.07029*, arXiv:2304.07029. https://doi.org/10.48550/arXiv.2304.07029

Chen, K., Han, T., Gong, J., Bai, L., Ling, F., Luo, J., Chen, X., Ma, L., Zhang, T., Su, R., Ci, Y., Li, B., Yang, X., & Ouyang, W. (2023). FengWu: Pushing the Skillful Global Medium-range Weather Forecast beyond 10 Days Lead. *arXiv e-prints*, arXiv:2304.02948 https://doi.org/10.48550/arXiv.2304.02948

Chen, L., Zhong, X., Zhang, F., Cheng, Y., Xu, Y., Qi, Y., & Li, H. (2023). FuXi: a cascade machine learning forecasting system for 15-day global weather forecast. *npj Climate and Atmospheric Science*, *6*(1), 190. https://doi.org/10.1038/s41612-023-00512-1

Darcet, T., Oquab, M., Mairal, J., & Bojanowski, P. (2024). Vision Transformers Need Registers. The Twelfth International Conference on Learning Representations,

Enström, D., Kjellberg, V., & Johansson, M. (2024). Reasoning in Transformers - Mitigating Spurious Correlations and Reasoning Shortcuts. *arXiv e-prints*, *abs/2403.11314*, arXiv:2403.11314. https://doi.org/10.48550/arXiv.2403.11314

Ghosal, S. S., & Li, Y. (2024). Are Vision Transformers Robust to Spurious Correlations? *International Journal of Computer Vision*, *132*(3), 689-709. https://doi.org/10.1007/s11263-023-01916-5

Giraldo, F. X. (2001). A spectral element shallow water model on spherical geodesic grids. *International Journal for Numerical Methods in Fluids*, *35*, 869-901.

Ha, S., & Lyu, I. (2022). SPHARM-Net: Spherical Harmonics-Based Convolution for Cortical Parcellation. *IEEE Transactions on Medical Imaging*, *41*(10), 2739-2751. https://doi.org/10.1109/TMI.2022.3168670

Han, D., Pan, X., Han, Y., Song, S., & Huang, G. (2023, 1-6 Oct. 2023). FLatten Transformer: Vision Transformer using Focused Linear Attention. 2023 IEEE/CVF International Conference on Computer Vision (ICCV),

He, K., Zhang, X., Ren, S., & Sun, J. (2016, 27-30 June 2016). Deep Residual Learning for Image Recognition. 2016 IEEE Conference on Computer Vision and Pattern Recognition (CVPR),

Hersbach, H., Bell, B., Berrisford, P., Hirahara, S., Horányi, A., Muñoz Sabater, J., Nicolas, J., Peubey, C., Radu, R., Schepers, D., Simmons, A., Soci, C., Abdalla, S., Abellan, X., Balsamo, G., Bechtold, P., Biavati, G., Bidlot, J., Bonavita, M., & Thépaut, J. N. (2020). The ERA5 global reanalysis. *Quarterly Journal of the Royal Meteorological Society*. https://doi.org/10.1002/qj.3803

Ho, J., Saharia, C., Chan, W., Fleet, D. J., Norouzi, M., & Salimans, T. (2022). Cascaded Diffusion Models for High Fidelity Image Generation. *J. Mach. Learn. Res.*, *23*, 47:41-47:33.

Hua, W., Dai, Z., Liu, H., & Le, Q. (2022). *Transformer Quality in Linear Time* Proceedings of the 39th International Conference on Machine Learning, Proceedings of Machine Learning Research. https://proceedings.mlr.press/v162/hua22a.html

John Xu, Z.-Q., Zhang, Y., Luo, T., Xiao, Y., & Ma, Z. (2020). Frequency Principle: Fourier Analysis Sheds Light on Deep Neural Networks. *Communications in Computational Physics*, *5*, 1746-1767. https://doi.org/http://doi.org/10.4208/cicp.OA-2020-0085

Kingma, D. P., & Ba, J. (2014). Adam: A Method for Stochastic Optimization. *arXiv e-prints*, *abs/1412.6980*, arXiv:1412.6980 https://doi.org/10.48550/arXiv.1412.6980

Koshyk, J. N., & Hamilton, K. (2001). The Horizontal Kinetic Energy Spectrum and Spectral Budget Simulated by a High-Resolution Troposphere–Stratosphere–Mesosphere GCM. *Journal of the Atmospheric Sciences*, *58*, 329-348.

Lai, T., Tran, Q. H., Bui, T., & Kihara, D. (2019). A Gated Self-attention Memory Network for Answer Selection. *Conference on Empirical Methods in Natural Language Processing*.

Lam, R., Sanchez-Gonzalez, A., Willson, M., Wirnsberger, P., Fortunato, M., Alet, F., Ravuri, S., Ewalds, T., Eaton-Rosen, Z., Hu, W., Merose, A., Hoyer, S., Holland, G., Vinyals, O., Stott, J., Pritzel, A., Mohamed, S., & Battaglia, P. (2023). Learning skillful medium-range global weather forecasting. *Science*, *382*(6677), 1416-1421. https://doi.org/10.1126/science.adi2336

Li, H., Lin, Z., Shen, X., Brandt, J., & Hua, G. (2015, 7-12 June 2015). A convolutional neural network cascade for face detection. 2015 IEEE Conference on Computer Vision and Pattern Recognition (CVPR),

Li, Z., Peng, J., & Zhang, L. (2023). Spectral Budget of Rotational and Divergent Kinetic Energy in Global Analyses. *Journal of the Atmospheric Sciences*, *80*(3), 813-831. https://doi.org/https://doi.org/10.1175/JAS-D-21-0332.1

Li, Z., Sun, T., Wang, H., & Wang, B. (2021). Adaptive and Implicit Regularization for Matrix Completion. *SIAM J. Imaging Sci.*, *15*, 2000-2022.

Li, Z., Wang, H., & Meng, D. (2023). Regularize implicit neural representation by itself. *2023 IEEE/CVF Conference on Computer Vision and Pattern Recognition (CVPR)*, 10280-10288.

Loshchilov, I., & Hutter, F. (2017a). Decoupled Weight Decay Regularization. International Conference on Learning Representations,

Loshchilov, I., & Hutter, F. (2017b). SGDR: Stochastic Gradient Descent with Warm Restarts. International Conference on Learning Representations,

Ma, X., Zhou, C., Kong, X., He, J., Gui, L., Neubig, G., May, J., & Zettlemoyer, L. (2023). Mega: Moving Average Equipped Gated Attention. The Eleventh International Conference on Learning Representations,

McCabe, M., Harrington, P., Subramanian, S., & Brown, J. (2023). Towards Stability of Autoregressive Neural Operators. *Transactions on Machine Learning Research*. https://openreview.net/forum?id=RFfUUtKYOG

Nguyen, T., Brandstetter, J., Kapoor, A., Gupta, J. K., & Grover, A. (2023). *ClimaX: A foundation model for weather and climate* Proceedings of the 40th International Conference on Machine Learning, Proceedings of Machine Learning Research. https://proceedings.mlr.press/v202/nguyen23a.html

Niranjan Kumar, K., Ashrit, R., Sreevathsa, R., Kumar, S., Mishra, A. K., Thota, M. S., Jayakumar, A., Mohandas, S., & Mitra, A. K. (2023). Atmospheric kinetic energy spectra from global and regional NCMRWF unified modelling system. *Quarterly Journal of the Royal Meteorological Society*, *149*(756), 2784-2799. https://doi.org/https://doi.org/10.1002/qj.4531

Paszke, A., Gross, S., Chintala, S., Chanan, G., Yang, E., Devito, Z., Lin, Z., Desmaison, A., Antiga, L., & Lerer, A. (2017). Automatic differentiation in PyTorch.

Pathak, J., Subramanian, S., Harrington, P. Z., Raja, S., Chattopadhyay, A., Mardani, M., Kurth, T., Hall, D., Li, Z.-Y., Azizzadenesheli, K., Hassanzadeh, P., Kashinath, K., & Anandkumar, A. (2022). FourCastNet: A Global Data-driven High-resolution Weather Model using Adaptive Fourier Neural Operators. *arXiv e-prints*, arXiv:2202.11214. https://doi.org/10.48550/arXiv.2202.11214

Poulenard, A., & Guibas, L. J. (2021, 20-25 June 2021). A functional approach to rotation equivariant non-linearities for Tensor Field Networks. 2021 IEEE/CVF Conference on Computer Vision and Pattern Recognition (CVPR),

Rahaman, N., Baratin, A., Arpit, D., Draxler, F., Lin, M., Hamprecht, F., Bengio, Y., & Courville, A. (2019). *On the Spectral Bias of Neural Networks* Proceedings of the 36th International Conference on Machine Learning, Proceedings of Machine Learning Research. https://proceedings.mlr.press/v97/rahaman19a.html

Rasp, S., Dueben, P. D., Scher, S., Weyn, J. A., Mouatadid, S., & Thuerey, N. (2020). WeatherBench: A Benchmark Data Set for Data-Driven Weather Forecasting. *Journal of Advances in Modeling Earth Systems*, *12*(11), e2020MS002203. https://doi.org/https://doi.org/10.1029/2020MS002203

Rasp, S., & Thuerey, N. (2021). Data-Driven Medium-Range Weather Prediction With a Resnet Pretrained on Climate Simulations: A New Model for WeatherBench. *Journal of Advances in Modeling Earth Systems*, *13*(2), e2020MS002405. https://doi.org/https://doi.org/10.1029/2020MS002405

Ronneberger, O., Fischer, P., & Brox, T. (2015). U-Net: Convolutional Networks for Biomedical Image Segmentation. *Lncs*, *9351*, 234-241. https://doi.org/10.1007/978-3-319-24574-4_28

Shen, Z., Shen, T., Lin, Z., & Ma, J. (2021). PDO-eS 2 CNNs: Partial Differential Operator Based Equivariant Spherical CNNs. *Proceedings of the AAAI Conference on Artificial Intelligence*, *35*, 9585-9593. https://doi.org/10.1609/aaai.v35i11.17154

Shi, Y., Sun, M., Wang, Y., Wang, R., Sun, H., & Chen, Z. (2024). FViT: A Focal Vision Transformer with Gabor Filter. *arXiv e-prints*, arXiv:2402.11303. https://doi.org/10.48550/arXiv.2402.11303

Ulyanov, D., Vedaldi, A., & Lempitsky, V. S. (2016). Instance Normalization: The Missing Ingredient for Fast Stylization. *arXiv e-prints*, arXiv:1607.08022. https://doi.org/10.48550/arXiv.1607.08022

Vaswani, A., Shazeer, N., Parmar, N., Uszkoreit, J., Jones, L., Gomez, A. N., Kaiser, Ł., & Polosukhin, I. (2017). *Attention is all you need* Proceedings of the 31st International Conference on Neural Information Processing Systems, Long Beach, California, USA.

Weyn, J., Durran, D., & Caruana, R. (2020). Improving Data-Driven Global Weather Prediction Using Deep Convolutional Neural Networks on a Cubed Sphere. *Journal of Advances in Modeling Earth Systems*, *12*. https://doi.org/10.1029/2020MS002109

Xu, W., & Wan, Y. (2024). ELA: Efficient Local Attention for Deep Convolutional Neural Networks. *arXiv e-prints*, arXiv:2403.01123. https://doi.org/10.48550/arXiv.2403.01123

You, H., Xiong, Y., Dai, X., Wu, B., Zhang, P., Fan, H., Vajda, P., & Lin, Y. C. (2023, 17-24 June 2023). Castling-ViT: Compressing Self-Attention via Switching Towards Linear-Angular Attention at Vision Transformer Inference. 2023 IEEE/CVF Conference on Computer Vision and Pattern Recognition (CVPR).

**Supplementary Information for "Long-Term Prediction Accuracy Improvement of Data-Driven Medium-Range Global Weather Forecast"**

Yifan Hu[1, 2], Fukang Yin[2, *], Weimin Zhang[2, *], Kaijun Ren[2], Junqiang Song[2], Kefeng Deng[2], and Di Zhang[2]

[1]College of Computer Science and Technology, National University of Defense Technology, Changsha, P.R. China, 410073
[2]College of Meteorology and Oceanography, National University of Defense Technology, Changsha, P.R. China, 410073

## Contents of this file

Supplementary Texts

Supplementary Tables S1 to S3

Supplementary Figures S1 to S11

**Supplementary Texts**

The kinetic energy (KE) spectra for different iteration steps are calculate as follows (Koshyk & Hamilton, 2001; Li et al., 2023; Niranjan Kumar et al., 2023):

$$U(\lambda, \varphi, p, t) = \sum_{n=0}^{N} \sum_{m=-n}^{n} U_n^m(p, t) P_n^m(cos\varphi) e^{im\lambda}$$

$$V(\lambda, \varphi, p, t) = \sum_{n=0}^{N} \sum_{m=-n}^{n} V_n^m(p, t) P_n^m(cos\varphi) e^{im\lambda}$$

$$E_n^m(u, v, p, t) = \frac{1}{2}(|U_n^m(p, t)|^2 + |V_n^m(p, t)|^2)$$

where $\lambda$ is longitude, $\varphi$ is latitude, $m$ is zonal wavenumber, $n$ is total wavenumber, $N$ denotes truncated wavenumber, $U_n^m$, $V_n^m$ are spectral coefficients of $U, V$, $P_n^m$ represents Legendre polynomials with degrees of freedom $n$, $E_n^m(u, v, p, t)$ is the KE spectra.

## References

Koshyk, J. N., & Hamilton, K. (2001). The Horizontal Kinetic Energy Spectrum and Spectral Budget Simulated by a High-Resolution Troposphere–Stratosphere–Mesosphere GCM. *Journal of the Atmospheric Sciences*, *58*, 329-348.

Li, Z., Peng, J., & Zhang, L. (2023). Spectral Budget of Rotational and Divergent Kinetic Energy in Global Analyses. *Journal of the Atmospheric Sciences*, *80*(3), 813-831. https://doi.org/https://doi.org/10.1175/JAS-D-21-0332.1

Niranjan Kumar, K., Ashrit, R., Sreevathsa, R., Kumar, S., Mishra, A. K., Thota, M. S., Jayakumar, A., Mohandas, S., & Mitra, A. K. (2023). Atmospheric kinetic energy spectra from global and regional NCMRWF unified modelling system. *Quarterly Journal of the Royal Meteorological Society*, *149*(756), 2784-2799. https://doi.org/https://doi.org/10.1002/qj.4531

## Supplementary Tables

**Table S1** Relative losses for the Shallow Water Equations on the rotating sphere at a spatial resolution of 64 × 128 and a temporal resolution of 1 hour. Lower is better. The reported lead times are 1h,10h and 50h respectively, which correspond to 1, 10 and 50 iteration steps.

| Model | 1h $L^2$ relative losses ($\times 10^2$) | | | 10h $L^2$ relative losses ($\times 10^2$) | | | 50h $L^2$ relative losses ($\times 10^2$) | | |
|---|---|---|---|---|---|---|---|---|---|
| | $Z$ | $U$ | $V$ | $Z$ | $U$ | $V$ | $Z$ | $U$ | $V$ |
| U-Net | 0.148 | 7.589 | 3.741 | 0.904 | 38.848 | 34.310 | 1.989 | 95.986 | 88.554 |
| FourCastNet | **0.061** | **2.648** | **2.274** | 0.171 | 8.476 | 7.013 | 0.393 | 25.136 | 25.703 |
| SFNO, Linear | 0.100 | 4.312 | 3.746 | 0.184 | 10.190 | 8.440 | 0.381 | 23.114 | 19.304 |
| SFNO, Non-Linear | 0.102 | 4.383 | 3.759 | 0.180 | 9.839 | 8.271 | 0.365 | 20.158 | 18.672 |
| SHNO-SMHSA | 0.131 | 5.765 | 5.296 | 0.244 | 14.827 | 13.009 | 0.537 | 38.818 | 38.841 |
| SHNO | 0.094 | 3.835 | 3.391 | **0.157** | **7.488** | **6.362** | **0.286** | **13.860** | **13.333** |

**Table S2** Model parameters for medium-range global weather forecast.

| Model | Parameters | | |
|---|---|---|---|
| | Layers | Embed. dimension | Parameter count |
| FourCastNet | 8 | 768 | 22.311M |
| SFNO | 4 | 928 | 21.066M |
| SHNO | 4 | 512 | 20.654M |

**Table S3** The abbreviations and their descriptions for different variables.

| Abbreviation | Description |
|---|---|
| 10U | zonal wind velocity at 10m from the surface |
| 10V | meridional wind velocity at 10m from the surface |
| T2M | temperature at 2m from the surface |
| U--- | zonal wind velocity at pressure level --- |
| V--- | meridional wind velocity at pressure level --- |
| Z--- | geopotential at pressure level --- |
| T--- | temperature at pressure level --- |
| RH--- | relative humidity at pressure level --- |

## Supplementary Figures

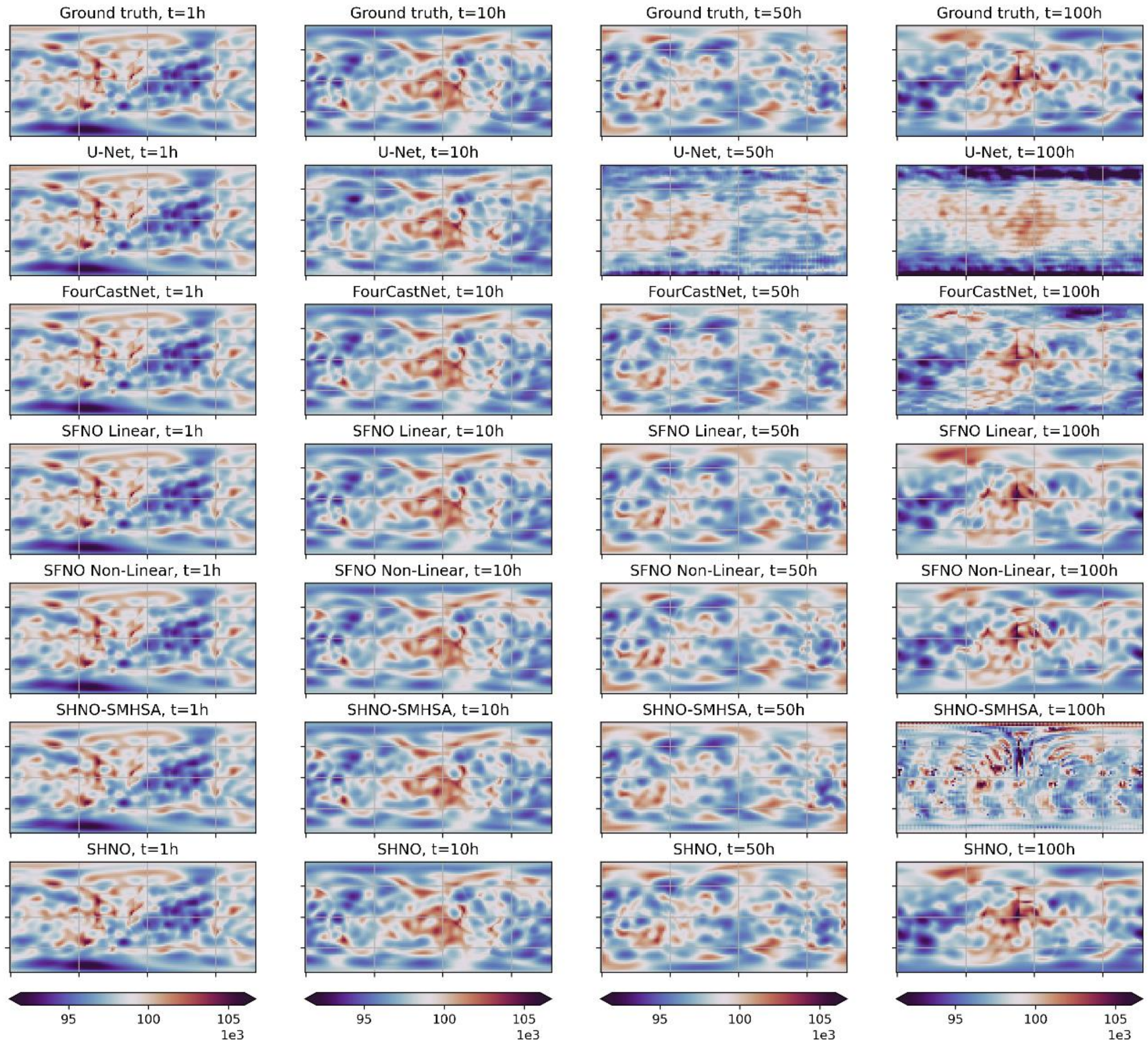

**Figure S1**: Visualization of forecast results for Spherical Shallow Water Equations. Columns from left to right corresponding to 1, 10, 50 and 100 iteration steps respectively. Rows from top to bottom represent the ground truth, U-Net, FourCastNet, SFNO Linear, SFNO Non-Linear, SHNO-SMHSA, and SHNO respectively. The initial input fields are the same as in Figure 2.

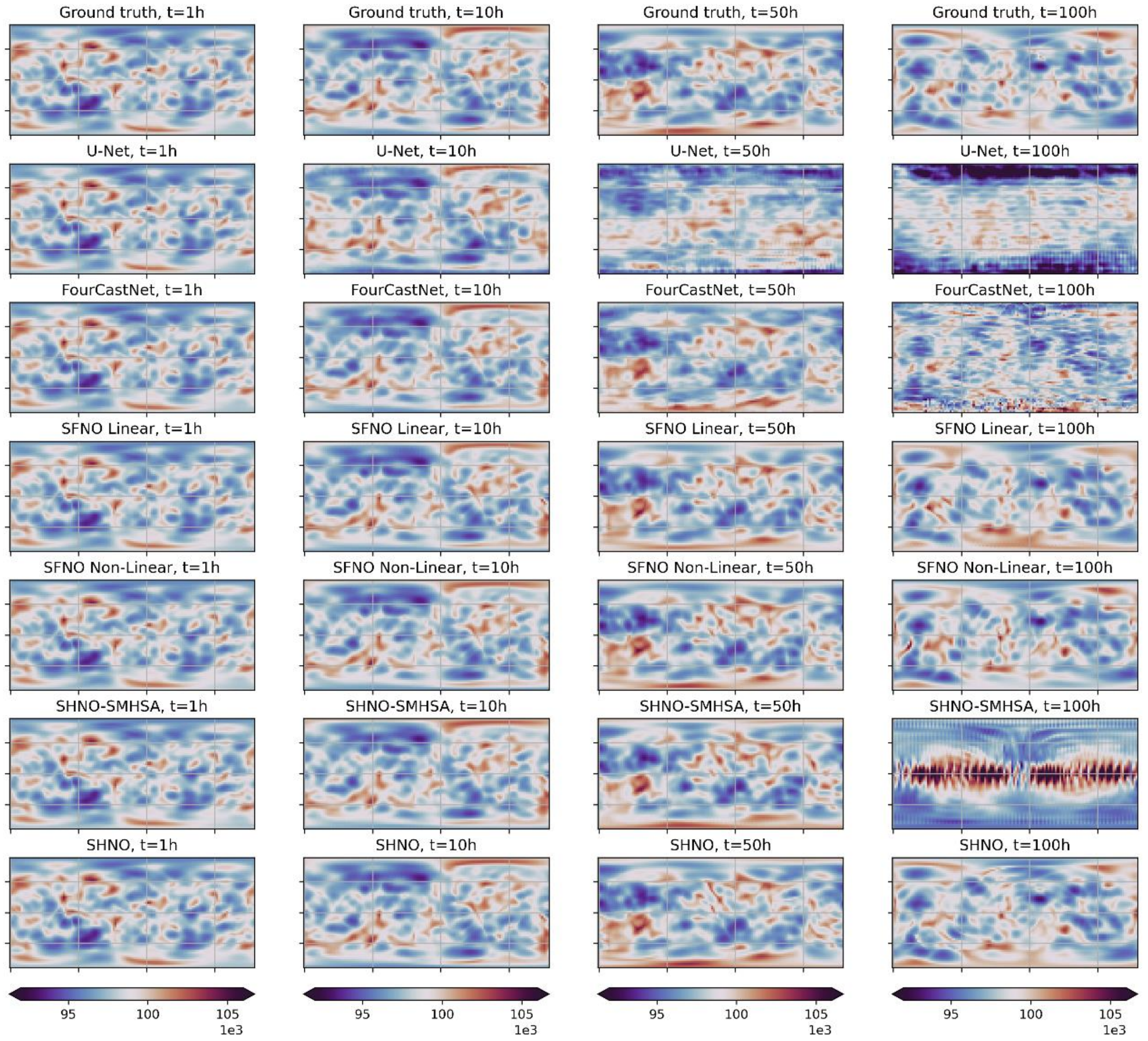

**Figure S2**: Visualization of forecast results for Spherical Shallow Water Equations. Columns from left to right corresponding to 1, 10, 50 and 100 iteration steps respectively. Rows from top to bottom represent the ground truth, U-Net, FourCastNet, SFNO Linear, SFNO Non-Linear, SHNO-SMHSA, and SHNO respectively. The initial input fields are the same as in Figure S3.

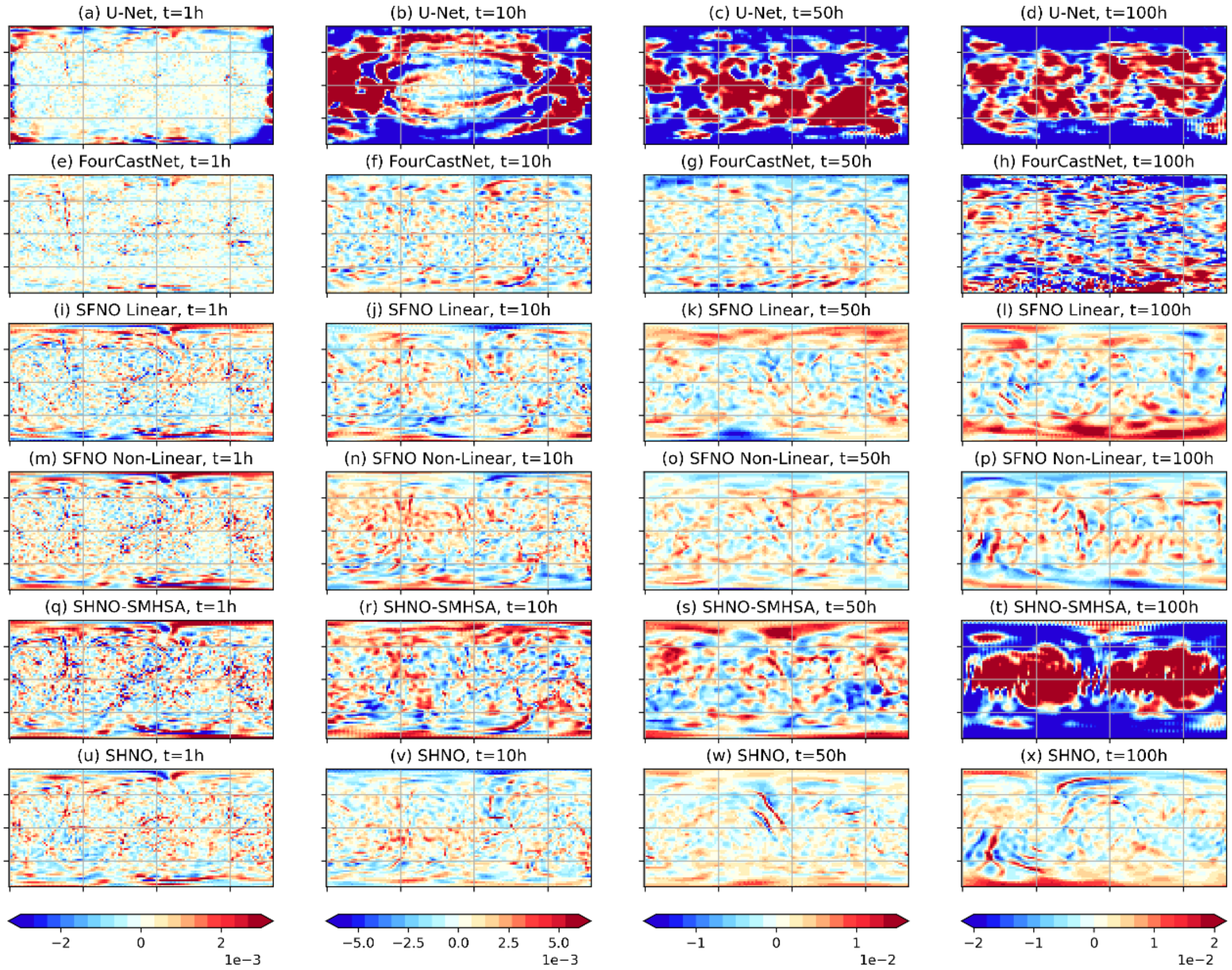

**Figure S3**: Spatial distribution of relative errors for geopotential height in Spherical Shallow Water Equations. The smaller the absolute value, the better the performance. Columns from left to right corresponding to 1, 10, 50 and 100 iteration steps respectively. Rows from top to bottom represent the ground truth, U-Net, FourCastNet, SFNO Linear, SFNO Non-Linear, SHNO-SMHSA, and SHNO respectively. The initial input fields are the same as in Figure S2.

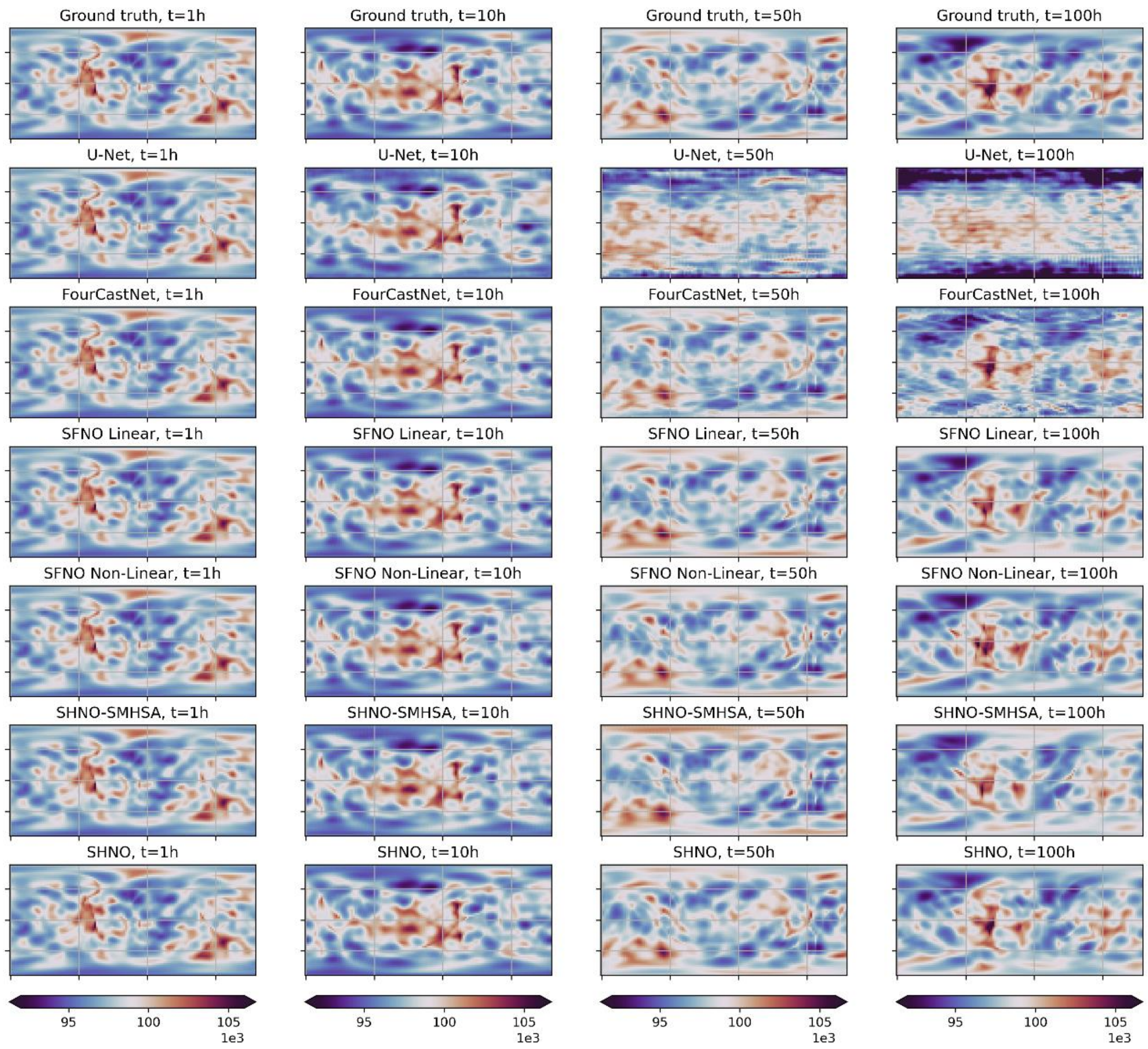

**Figure S4**: Visualization of forecast results for Spherical Shallow Water Equations. Columns from left to right corresponding to 1, 10, 50 and 100 iteration steps respectively. Rows from top to bottom represent the ground truth, U-Net, FourCastNet, SFNO Linear, SFNO Non-Linear, SHNO-SMHSA, and SHNO respectively. The initial input fields are the same as in Figure S5.

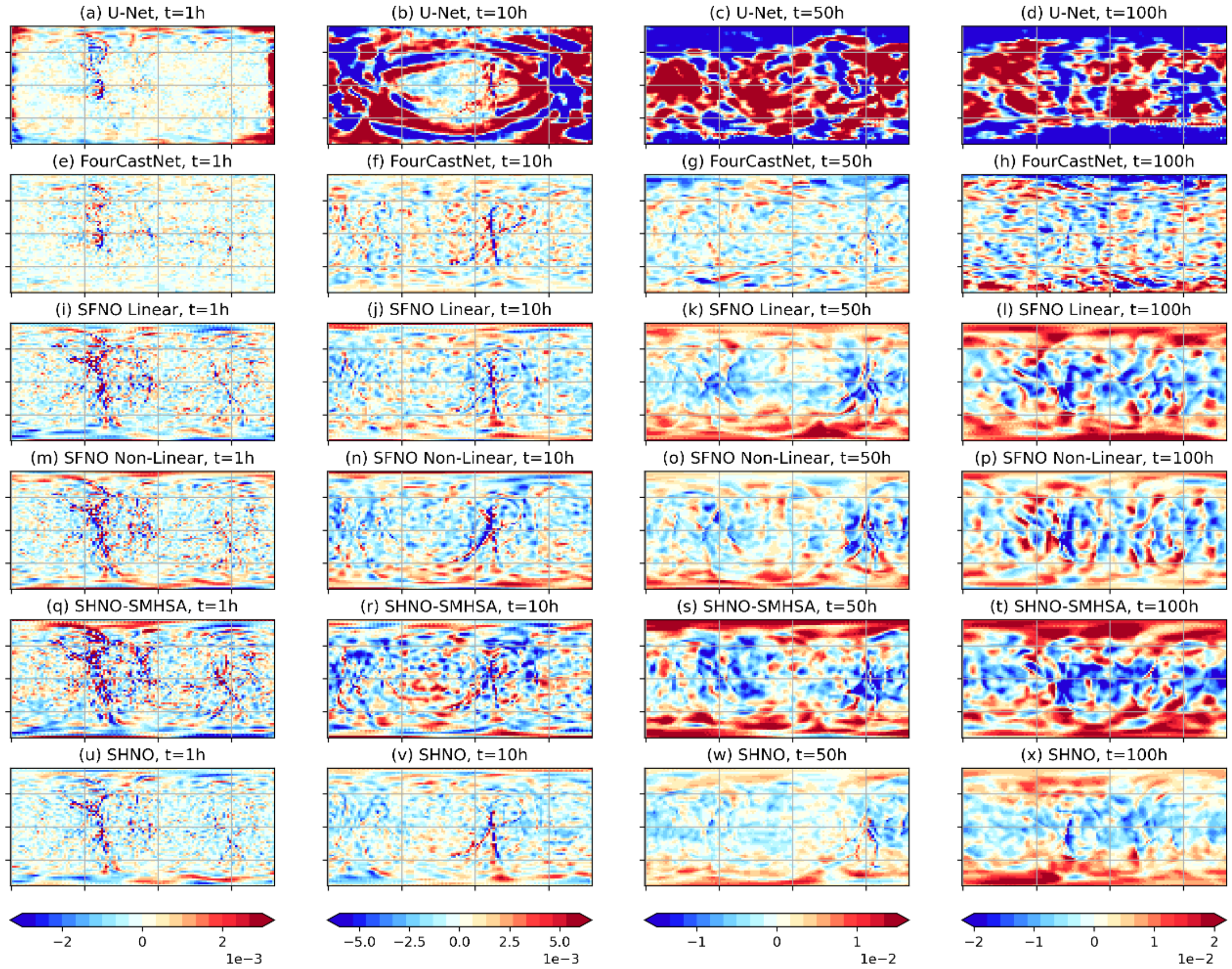

**Figure S5**: Spatial distribution of relative errors for geopotential height in Spherical Shallow Water Equations. The smaller the absolute value, the better the performance. Columns from left to right corresponding to 1, 10, 50 and 100 iteration steps respectively. Rows from top to bottom represent the ground truth, U-Net, FourCastNet, SFNO Linear, SFNO Non-Linear, SHNO-SMHSA, and SHNO respectively. The initial input fields are the same as in Figure S4.

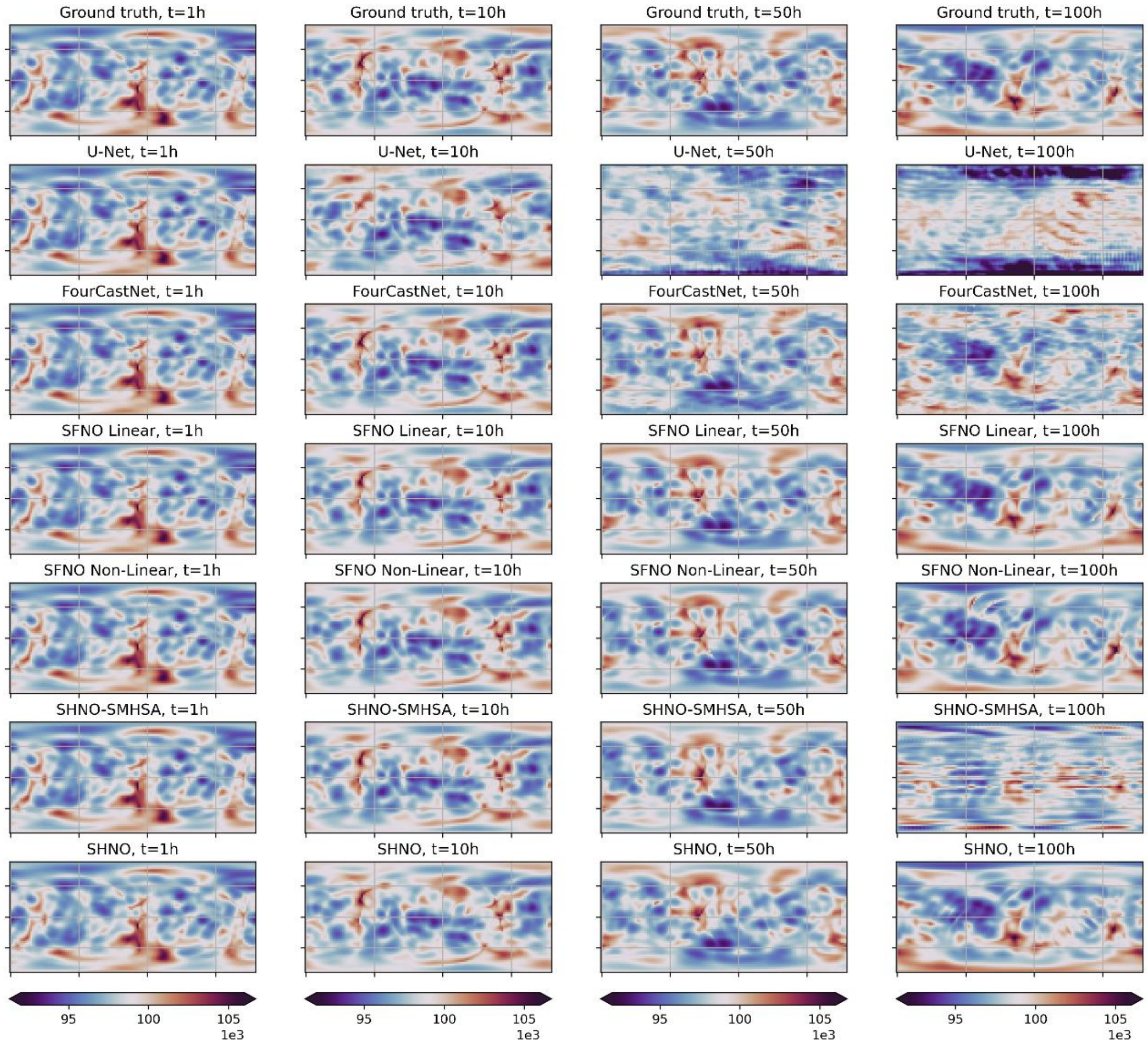

**Figure S6**: Visualization of forecast results for Spherical Shallow Water Equations. Columns from left to right corresponding to 1, 10, 50 and 100 iteration steps respectively. Rows from top to bottom represent the ground truth, U-Net, FourCastNet, SFNO Linear, SFNO Non-Linear, SHNO-SMHSA, and SHNO respectively. The initial input fields are the same as in Figure S7.

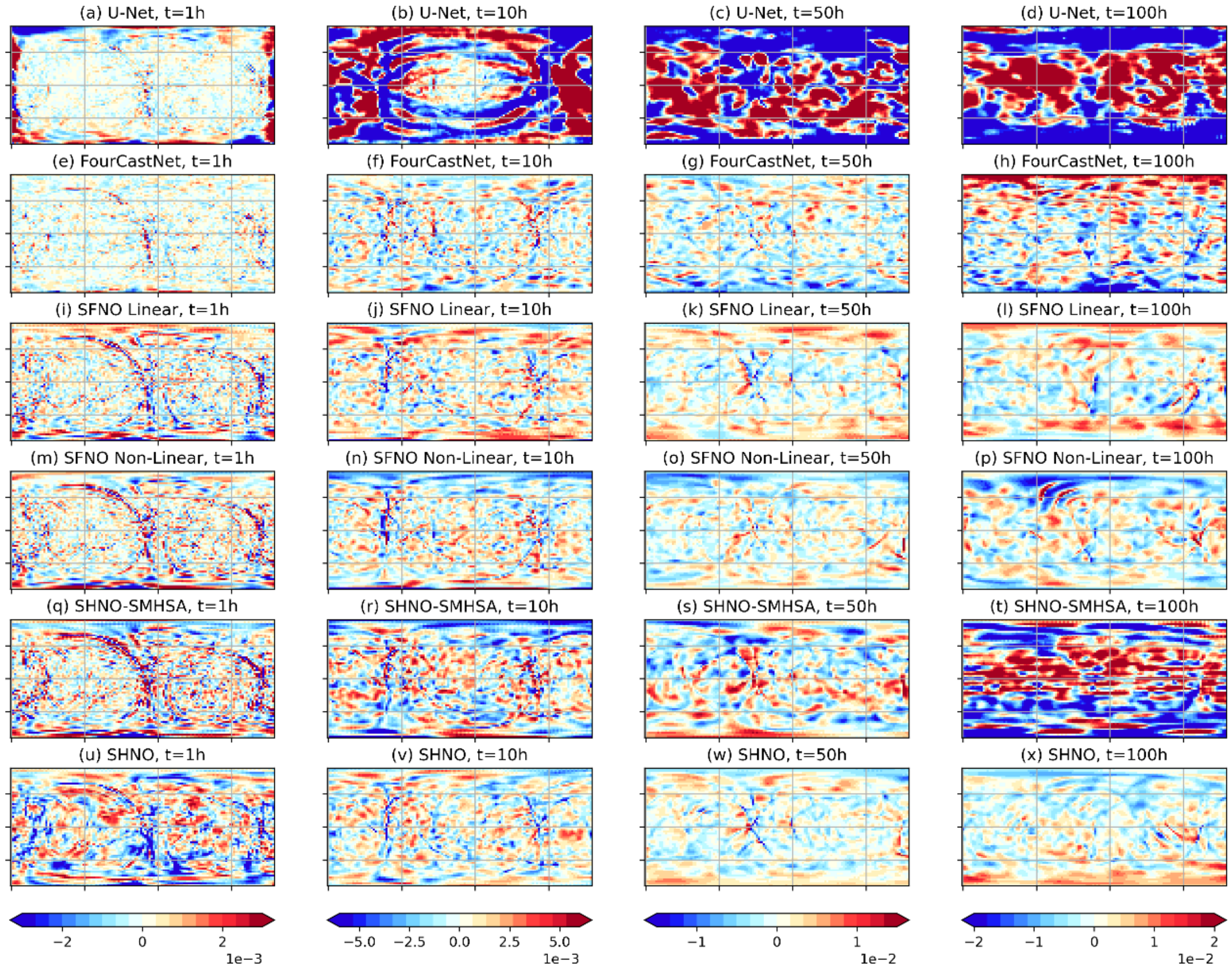

**Figure S7**: Spatial distribution of relative errors for geopotential height in Spherical Shallow Water Equations solving. The smaller the absolute value, the better the performance. Columns from left to right corresponding to 1, 10, 50 and 100 iteration steps respectively. Rows from top to bottom represent the ground truth, U-Net, FourCastNet, SFNO Linear, SFNO Non-Linear, SHNO-SMHSA, and SHNO respectively. The initial input fields are the same as in Figure S6.

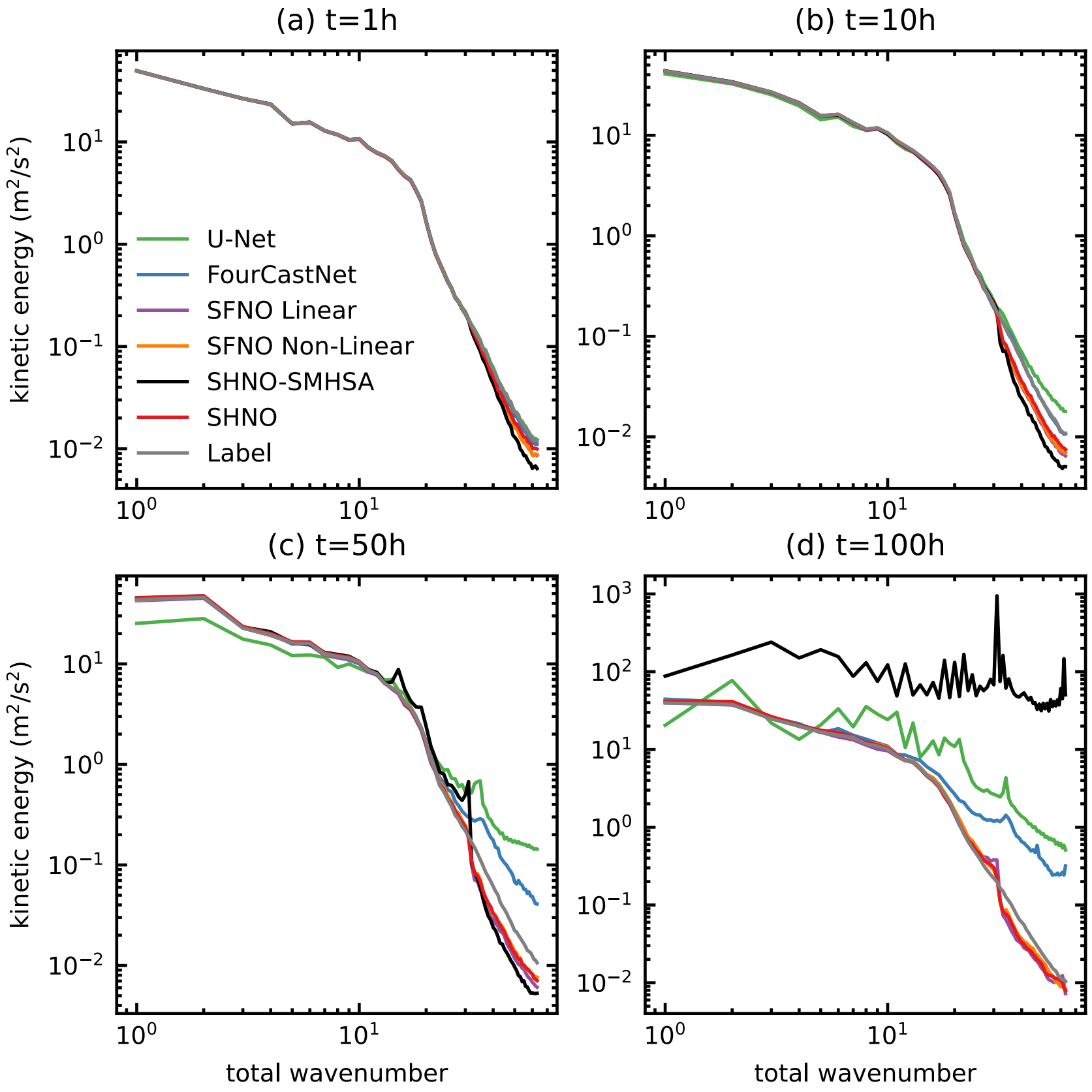

**Figure S8**: The kinetic energy spectra for different iteration steps i.e. lead times. The sign t in the figure represents the lead time: (a) 1 iteration step i.e. the lead time is 1h; (b) 10 iteration steps i.e. the lead time is 10h; (c) 50 iteration steps i.e. the lead time is 50h; (d) 100 iteration steps i.e. the lead time is 100h.

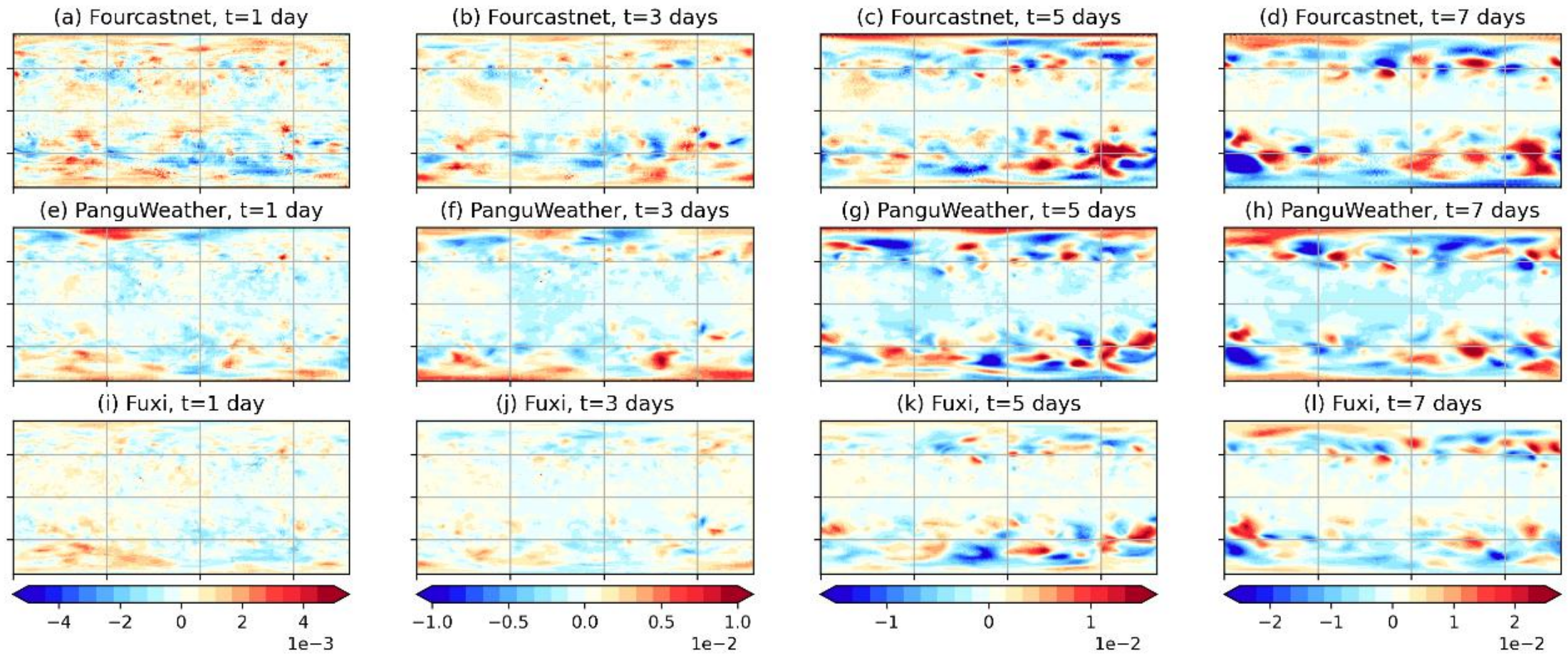

**Figure S9**: Spatial distribution of relative errors for geopotential height at 250hPa pressure level. Columns from left to right correspond to 1 day, 3 days, 5 days, and 7 days lead time i.e. 4, 12, 20, and 28 iteration steps respectively. Rows from top to bottom represent the Fourcastnet, PanguWeather, and Fuxi models. For all cases, the input time is 00:00 UTC on 1 September 2018, and the spatial resolution is 0.25° × 0.25°.

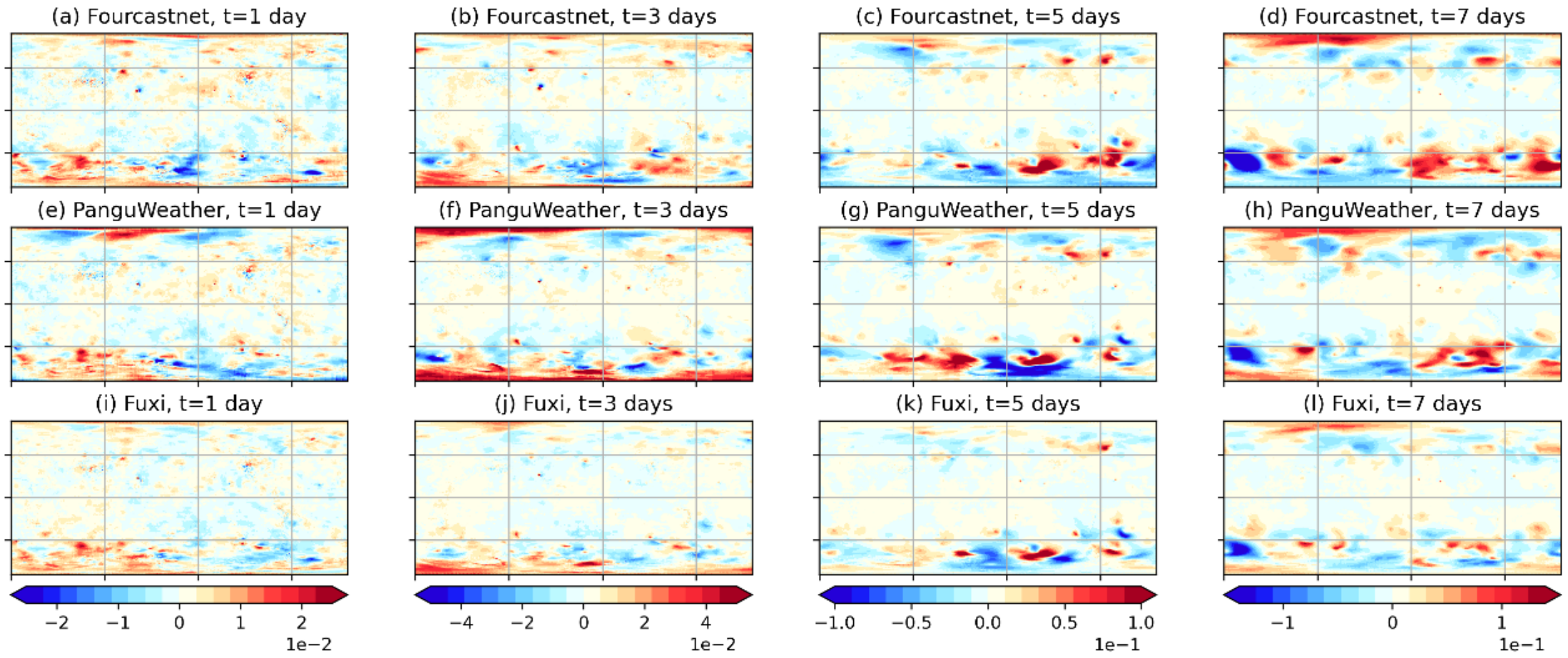

**Figure S10**: Spatial distribution of relative errors for geopotential height at 850hPa pressure level. Columns from left to right correspond to 1 day, 3 days, 5 days, and 7 days lead time i.e. 4, 12, 20, and 28 iteration steps respectively. Rows from top to bottom represent the Fourcastnet, PanguWeather, and Fuxi models. For all cases, the input time is 00:00 UTC on 1 September 2018, and the spatial resolution is 0.25° × 0.25°.

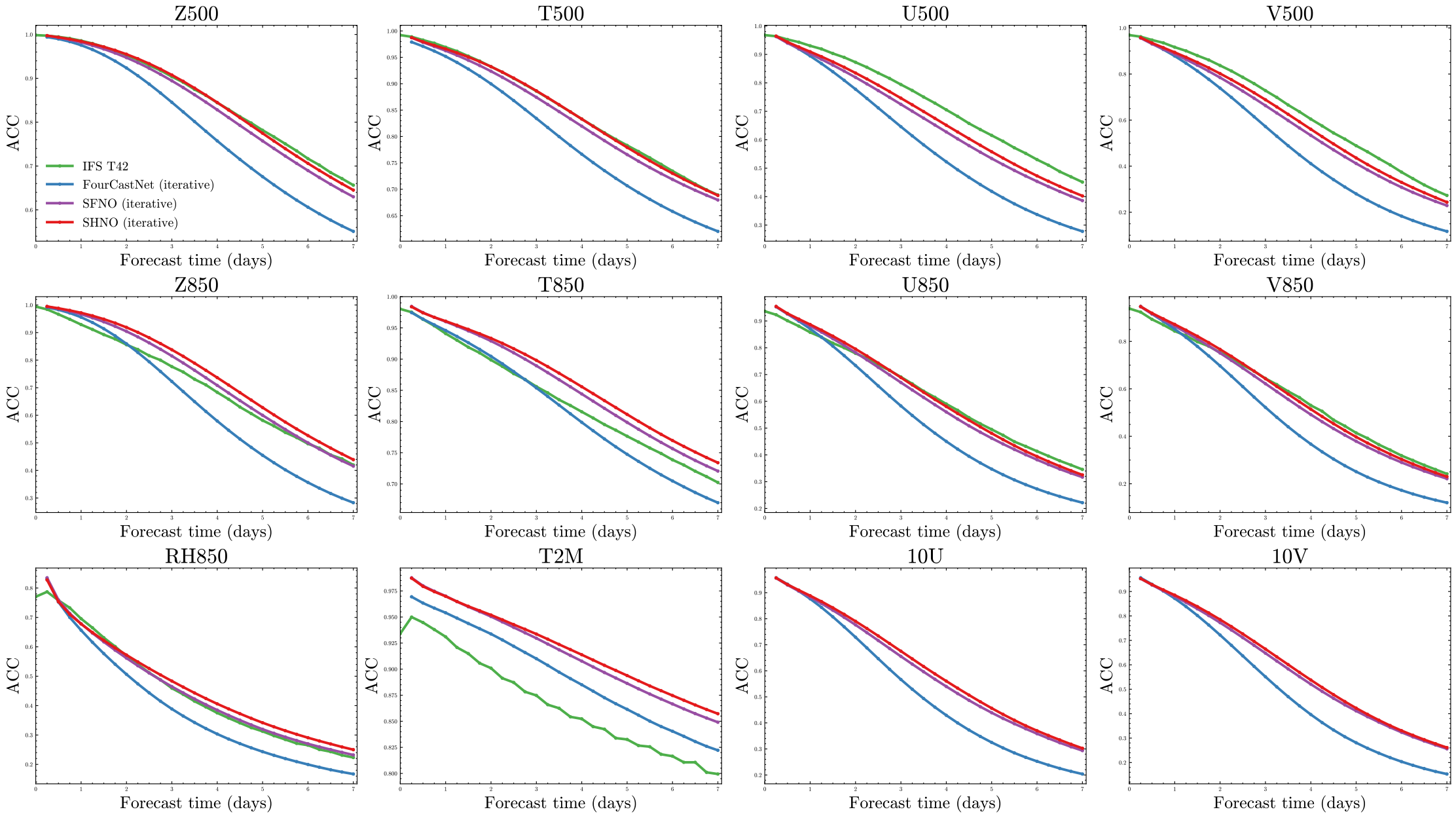

**Figure S11**: Globally averaged latitude-weighted ACC of IFS T42 (green line), FourCastNet (blue lines), SFNO (purple lines), and SHNO (red lines) for 3 surface variables, 4 upper-air variables at 500hPa pressure level, and 5 upper-air variables at 850hPa pressure level with spatial resolution of 5.625° × 5.625° in 7 days forecasts using testing data from 2017 to 2018. Higher is better. IFS T42 were not available for 10U and 10V.